\pdfoutput=1

\documentclass[11pt]{article}

\usepackage[final]{acl}

\usepackage{times}
\usepackage{latexsym}
\usepackage{array}
\usepackage[T1]{fontenc}

\usepackage[utf8]{inputenc}

\usepackage{microtype}

\usepackage{inconsolata}

\usepackage{graphicx}

\usepackage{amsmath}
\usepackage{amsfonts}
\usepackage{algorithm}
\usepackage{algpseudocode}
\usepackage{xcolor}
\usepackage{mdwlist}
\usepackage{multirow}
\usepackage{subcaption}
\usepackage{float}
\usepackage{epigraph}
\usepackage[table]{xcolor}
\usepackage{booktabs}
\usepackage{pifont}
\newcommand{\cmark}{\ding{51}}%
\newcommand{\xmark}{\ding{55}}%

\title{Meta-Pretraining for Zero-Shot Cross-Lingual Named Entity Recognition in Low-Resource Philippine Languages}

\author{
    \textbf{
        David Demitri Africa\thanks{Corresponding author: david.demitri.africa@gmail.com} ~~~
        Suchir Salhan ~~~
        Yuval Weiss
        } \\
    \textbf{
         Paula Buttery
         ~~~
         Richard Diehl Martinez
         }\\
    University of Cambridge
}

\begin{document}
\maketitle
\begin{abstract}
Named-entity recognition (NER) in low-resource languages is usually tackled by finetuning very large multilingual LMs, an option that is often infeasible in memory- or latency-constrained settings.  We ask whether small decoder LMs can be pretrained so that they adapt quickly and transfer zero-shot to languages unseen during pretraining.  To this end we replace part of the autoregressive objective with first-order model-agnostic meta-learning (MAML).  Tagalog and Cebuano are  typologically similar yet structurally different in their actor/non-actor voice systems, and hence serve as a challenging test-bed.  Across four model sizes (11 M – 570 M) MAML lifts zero-shot micro-F\textsubscript{1} by 2–6 pp under head-only tuning and 1–3 pp after full tuning, while cutting convergence time by up to 8\%.  Gains are largest for single-token person entities that co-occur with Tagalog case particles \emph{si}/\emph{ni}, highlighting the importance of surface anchors. 
\end{abstract}

\section{Introduction}

Named-entity recognition (NER) locates and categorises Persons (\texttt{PER}), Organisations (\texttt{ORG}) and Locations (\texttt{LOC}) in unstructured text \citep{chinchor1997muc}. It is used in a variety of important domains such as healthcare \citep{kundeti2016clinical, polignano2021comparing, shafqat2022standard} and law \citep{leitner2019fine,au2022ner, naik2023legal}, yet progress remains concentrated in a handful of well-resourced languages. Cross-lingual named-entity recognition is therefore important to better serve underserved communities, yet recent advancements remain unevenly distributed since NER performance in many languages remains poor due to limited training resources.

A key challenge is that entity boundaries and categories are not universal: languages differ in their morphosyntactic cues, word order, and orthographic conventions. Models trained primarily on Indo-European data thus fail to generalize reliably to underrepresented settings. In this paper, we address this problem through \textbf{meta-pretraining}: shaping language model initializations to adapt rapidly to new linguistic conditions. Unlike standard pretraining, which minimizes average loss over a static corpus, episodic meta-pretraining (e.g.\ via MAML; \citealt{pmlr-v70-finn17a}) explicitly optimizes for fast transfer. For low-resource NER, this offers two potential benefits: (i) rapid adaptation to languages with typologically distinct cues (e.g.\ case particles, voice systems, code-switching), and (ii) stronger zero-shot prototypes for common entity types, even without in-language exposure. While meta-learning has been explored for classification tasks in English or cross-lingually at BERT scale \citep{wu2020enhanced, li2020few, de2021meta}, its efficacy for small decoder LMs and morphologically rich languages is underexplored.

\begin{table}[t]
\centering
\small
\resizebox{\linewidth}{!}{
\begin{tabular}{l p{2.8cm} p{2.8cm}}
\toprule
\textbf{Typological Feature} & \textbf{Tagalog} & \textbf{Cebuano} \\
\midrule
Voice system            & \cellcolor{orange!30}\cmark\ Four-way & \cellcolor{orange!30}\cmark\ Reduced two-way \\
Case marking            & \cellcolor{orange!30}\cmark\ Obligatory & \cellcolor{orange!30}\xmark\ Often dropped \\
Borrowing / code-switch & \cellcolor{yellow!30}\cmark\ High density & \cellcolor{yellow!30}\xmark\ More conservative \\
Morphological richness  & \cellcolor{orange!30}\cmark\ Productive affixation & \cellcolor{orange!30}\cmark\ Regular affixation \\
Word order flexibility  & \cellcolor{green!30} \cmark\  & \cellcolor{green!30} \cmark\  \\
Pronominal systems      & \cellcolor{green!30} \cmark\ Rich clitic pronouns &  \cellcolor{green!30} \cmark\ Similar \\
Reduplication           & \cellcolor{orange!30}\cmark\ Common & \cellcolor{orange!30}\cmark\ Widespread \\
Orthography variation   & \cellcolor{yellow!30}\cmark\ Multiple conventions & \cellcolor{yellow!30}\xmark\ Multiple conventions \\
Pivot marking           & \cellcolor{orange!30}\cmark\ Consistently overt & \cellcolor{orange!30}\cmark\ Overt but less consistent \\
\bottomrule
\end{tabular}}
\caption{A selection of Typological Features of Tagalog and Cebuano relevant for NER. \cmark\ indicates strong presence, \xmark\ indicates reduced/less overt presence in each language. We highlight \textcolor{orange!80!black}{\textbf{high divergence}} features,  \textcolor{yellow!70!black}{\textbf{moderate divergence}} and \textcolor{green!70!black}{\textbf{similar}}  features compared to Indo-European Languages, motivating these languages as a case-study for low-resourced NER. We provide a more detailed comparison along with an illustrative gloss in Appendix \ref{app:linguistic}.}
\label{tab:tagalog-cebuano-summary}
\end{table}

As a case study, we focus on NER in Tagalog and Cebuano, the two most widely spoken Philippine languages \citep{miranda-2023-developing}. Typologically, both languages combine Austronesian features such as voice alternations, case particles, and reduplication with pervasive borrowing and code-switching (Figure~\ref{fig:intro-example}; Table~\ref{tab:tagalog-cebuano-summary}). These languages stress-test whether meta-pretraining can yield more adaptable NER representations than vanilla pretraining alone. We ask the following research questions: 
\begin{description}
  \item[RQ1] \textbf{Efficacy.} How much does first-order MAML improve zero-shot NER on Tagalog and Cebuano relative to vanilla autoregressive pretraining? 
  \item[RQ2] \textbf{What transfers?} Which entity classes, morphological cues, and lexical patterns (especially those tied to Tagalog/Cebuano typology) explain the observed gains or failures?
\end{description}

We answer these questions by systematically comparing first-order MAML and vanilla pretraining on LLaMa-style Pico Decoders across scales, analyzing both downstream performance and representation dynamics \citep{pico2025, pico}. This allows us to investigate:
\begin{description}
  \item[RQ3] How does the effect of meta-pretraining vary with model size? Are benefits stronger at small scales, or do they persist as capacity increases?
\end{description}

\subsection{Contributions.}
We provide the following contributions:
\begin{itemize}
  \item A systematic evaluation of meta-pretrained small decoder LMs for zero-shot NER in Tagalog and Cebuano, comparing against strong vanilla pretraining baselines across four model scales.
  \item Quantitative and qualitative evidence that MAML-based meta-pretraining produces sharper single-token entity prototypes, improving zero-shot NER, especially for person entities and Tagalog’s particle-rich syntax.
  \item An analysis of failure modes and learning dynamics, showing the capacity-dependent nature of meta-learning gains and the tradeoff between prototype sharpening and contextual generalization.
\end{itemize}

\section{Method}
\label{sec:method}

\subsection{Motivation}

\paragraph{Why these two languages?}
Tagalog and Cebuano are used every day by well over 100 million people. However, they occupy only a small fraction of the web text that current language models are pretrained on, which makes them both socially important and under-served by existing NLP tools \citep{miranda-2023-developing}. Linguistically, these languages also offer complementary typological challenges for NER, which we summarise in  Figure~\ref{tab:tagalog-cebuano-summary}. Tagalog and Cebuano combine Austronesian voice systems, case particles, reduplication, and discourse-driven topic marking in ways that are rare in widely studied NLP benchmarks. In particular, Tagalog offers more overt morphosyntactic cues than Cebuano: it retains  a four-way actor/non-actor voice paradigm, while Cebuano reduces this to two \citep{Tanangkingsing-2011-Cebuano} and  marks syntactic roles with case particles (\emph{si/ni/ang/ng/sa}). These languages offer a test bed for multilingual NER models that must generalize beyond Indo-European NER cues – where entities are typically identifiable through fixed word order and stable orthography– to handle the interaction of morphological marking, argument interaction and code-switching. Tagalog contains more Spanish loans and code-switching into English, while Cebuano maintains a more conservative Austronesian lexicon \citep{bautista2004tagalog, baklanova2019}. We provide a more detailed comparison of Tagalog and Cebuano typological features in Table~\ref{tab:tagalog-cebuano-appendix}.

\paragraph{Why Meta-learning?}
Being underrepresented in natural language processing (NLP) corpora \cite{cajote2024philippine, quakenbush2005philippine, dita2009building, bandarkar-etal-2024-belebele}, Philippine language datasets suffer from size and quality issues. In low-resource settings, where pretraining data is scarce or absent, it is important to ask the question: will a given checkpoint finetune or transfer rapidly when exposed to a novel language (such as in deployment)?

Meta-learning addresses this by shaping initializations for quick adaptation. Model-Agnostic Meta-Learning (MAML) optimizes an LM backbone so that a few gradient steps yield high performance on a new task \citep{pmlr-v70-finn17a}. We ask whether such an initialization, learned entirely without Tagalog/Cebuano exposure, can transfer to these languages’ distinct morphological and lexical cues for NER. Our working hypothesis is that a pretraining routine that is itself optimized for rapid adaptation will induce representations that generalize more readily across languages. Prior NLP studies have tested this mostly on English or on “BERT-scale” encoder models \citep{wu2020enhanced, ma2022decomposed, li2020few, de2021meta}; we explore whether episodic meta-pretraining of small decoder LMs, without any exposure to Tagalog or Cebuano, can still yield zero-shot gains for NER. We do not evaluate a multilingual language-model baseline, as our objective is to isolate the effect of episodic meta-pretraining under a matched corpus and schedule; training a competitive multilingual baseline would require different data and budgets, confounding a like-for-like comparison.

Our working hypothesis is that a pretraining routine that is itself
optimized for rapid adaptation will induce representations that
generalize more readily across languages, so that a model exposed only to high-resource sources can still zero-shot transfer to typologically
distant, low-resource targets.

\subsection{Architecture}

We build upon the \textsc{Pico} decoder stack \citep{pico2025}, a LLaMa-style causal Transformer implemented in PyTorch.  Four capacity tiers (\textbf{tiny} (11 M), \textbf{small} (65 M), \textbf{medium} (181 M) and \textbf{large} (570 M)) share all hyper-parameters except hidden width $d\!\in\!\{96,384,768,1536\}$.  Each model comprises $L{=}12$ RMS-normalised decoder blocks \citep{zhang2019rms} with grouped-query self-attention \citep{ainslie2023gqa}, RoPE positions \citep{su2024rope} and SwiGLU feed-forwards \citep{shazeer2020glu} that expand to $4d$.

\subsection{Hybrid pretraining objective}

Training alternates between two outer-loop updates:

\begin{enumerate}
    \item \textbf{Autoregressive LM step.}\; Standard next-token prediction on a pre-tokenized version of Dolma \cite{soldaini2024dolma} released by the Pico  library \cite{pico2025}.
    \item \textbf{First-order MAML episode.}\; A 32-way, 4-shot Subset-Masked LM Task (SMLMT; \citealp{bansal-etal-2020-self}) is sampled, where the model predicts a masked token from the corpus on the fly. 
          The inner loop finetunes a lightweight MLP head for ten SGD steps ($\alpha\!=\!10^{-3}$) and 
          the outer loop back-propagates the query loss through the frozen backbone.
\end{enumerate}

The branch decision is a Bernoulli draw with probability $\rho\!=\!0.5$, synchronised across four A100-80 GB GPUs. The pseudocode for both can be found in Appendix \ref{app:pseudocode}.

\subsection{Optimisation and monitoring}

We run $6{,}000$ outer updates with AdamW ($\eta_\text{peak}=3{\times}10^{-4}$, 2.5 k warm-up, cosine decay), accumulating eight micro-batches of 256 sequences to reach an effective batch of 2048 sequences (1024 for \textbf{tiny}).  Every 100 steps we log: Paloma perplexity \citep{magnusson2024paloma}, singular-value spectra of three attention and three feed-forward weight matrices, from which we compute proportional effective rank (PER; \citealp{martinez2024}), and support and query accuracy within MAML episodes.

\subsection{Finetuning on High-Resourced Languages}

We deliberately choose high-resource languages as the finetuning
sources because, in realistic deployments, these are the languages for which sizable, high-quality NER data already exists. They therefore form the most natural setting for cross-lingual transfer into low-resource settings.

After pretraining we attach an untrained linear conditional random field head \citep{10.5555/645530.655813}, which is a well-known method used often for NER \citep{bundschus2008extraction, ma-hovy-2016-end}. We finetune on a high-resource language (Danish, English, Croatian, Portuguese, Slovak, Serbian, Swedish, Chinese, Chinese-Simplified, and a mixture of all languages) before zero-shot evaluation on Tagalog (\texttt{tl\_trg}, \texttt{tl\_ugnayan}) and Cebuano (\texttt{ceb\_gja}) from Universal NER v1 \citep{mayhew-etal-2024-universal}. Results are later broken down by finetuning language. Further, two finetuning regimes are compared: head-only, where the transformer is frozen and only the classifier learns, and full, where all parameters are freed to update.

Finetuning uses AdamW ($3{\times}10^{-5}$) for up to ten epochs with early stopping on development F$_1$.  We report micro-F$_1$, with full details in Appendix \ref{app:uner}. 

\subsection{Baselines}

For each capacity tier we also evaluate a "vanilla" Pico model (no MAML, pure autoregressive loss) under identical data, schedule and compute. Pretraining results can be found in Appendix \ref{app:pretraining} with model configuration details in Appendix \ref{app:config}. A more detailed discussion of pretraining results and overall methodology can be found in \citet{africa2025learning}. 

\section{Zero-Shot Transfer Results}
\label{sec:filipino}

\noindent\textbf{Zero-shot evaluation.}  Unless stated otherwise, all scores are obtained without seeing any Tagalog/Cebuano data during finetune, relying solely on the UNER test sets (§ 2.4).

\begin{figure}[t]
  \centering
  \includegraphics[width=\linewidth]{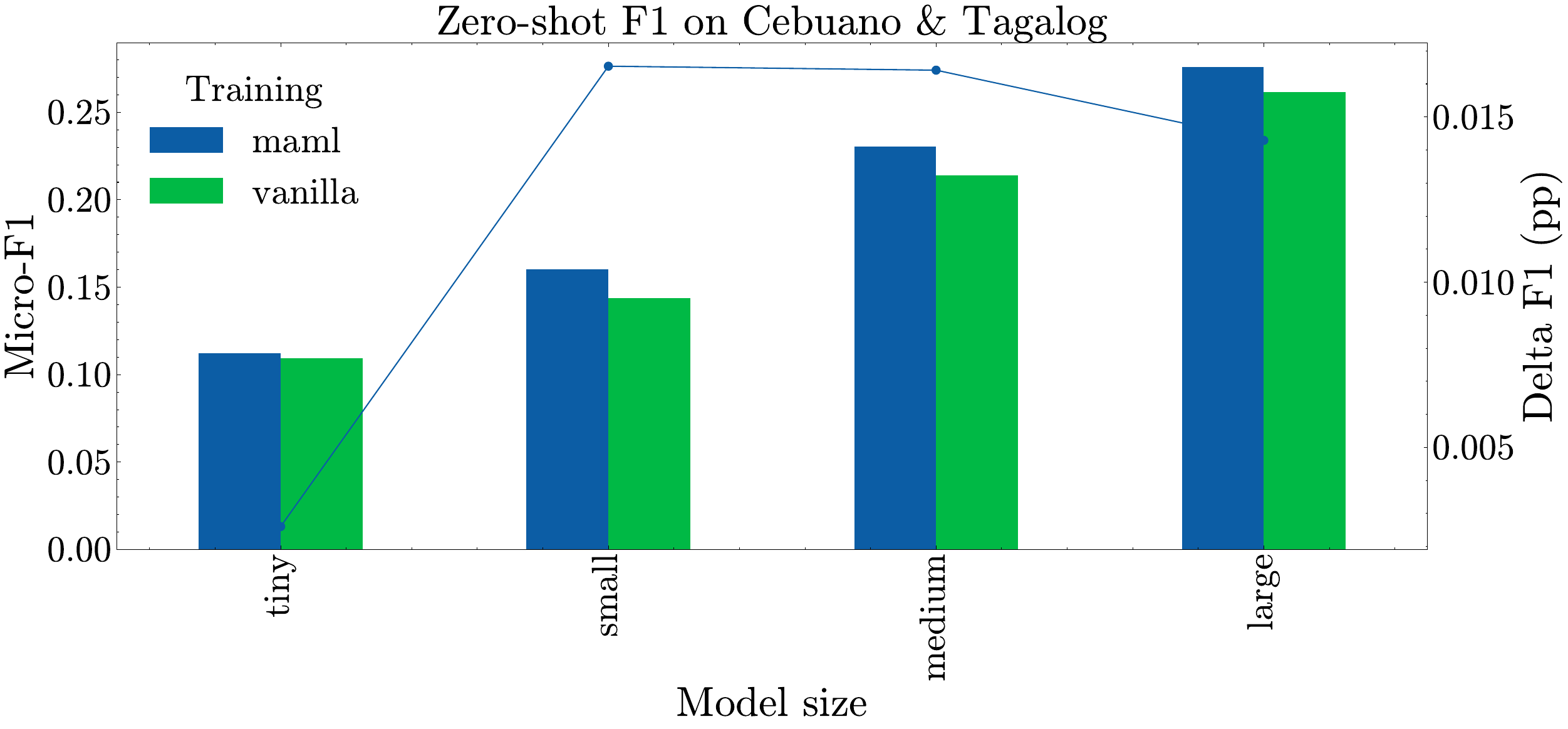}
  \caption{\textbf{Scale curve.}  
           Zero-shot Micro-F$_1$ on Cebuano \& Tagalog versus parameter count.
           Bars compare \textsc{Pico-MAML} (blue) to vanilla pretraining (green); the overlaid line shows the relative gain of MAML (Delta F1, right axis). Meta-pretraining helps at every scale, but the relative lift shrinks from +38 \% (11 M) to +6 \% (570 M), revealing a capacity threshold below which the inner loop cannot extract reusable features.}
  \label{fig:zero_shot_by_size}
\end{figure}

Figure \ref{fig:zero_shot_by_size} shows that \textsc{Pico-MAML} improves Cebuano/Tagalog micro-F$_1$ at every parameter budget. The relative lift is largest for moderate sizes and tapers with scale (+6\% at 570M). These results indicate that adding a single outer-loop meta-update per batch yields a cross-lingual prior not captured by vanilla pretraining under our setup.

\begin{figure}[t]
  \centering
  \includegraphics[width=\linewidth]{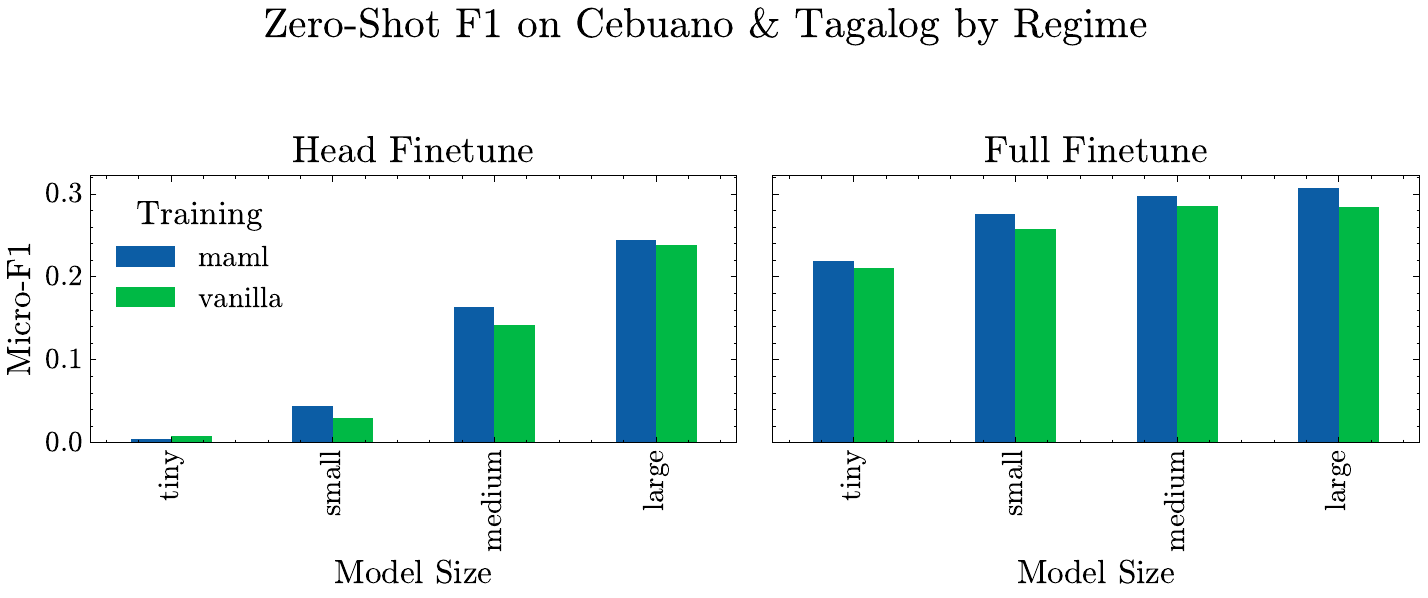}
  \caption{\textbf{Impact of finetuning regime.}  
           Head-only tuning (left) magnifies the meta-learning advantage up to +2.5 pp at 570 M, likely because the backbone must already encode entity cues. Full tuning (right) reduces but does not erase the gap, suggesting that MAML primarily accelerates convergence rather than acting as a regulariser.}  \label{fig:zero_shot_by_regime}
\end{figure}

\paragraph{Comparison of head-only tuning and full tuning.} Decomposing by finetuning regime (Fig.\;\ref{fig:zero_shot_by_regime}), MAML yields 1–2 pp gains when only the CRF head is trained, implying that the frozen weights already embeds better entity cues. Full tuning narrows the gap to 0.5–1.3 pp, indicating that the lift persists even when the optimiser is free to overwrite the initialisation.

Further, results indicate that the benefit provided by the meta-objective is scale-dependent. For the 11 M (\textbf{tiny}) model, MAML moves the overall score by $<1$ pp and yields no gain under head-only tuning.  From 65 M parameters upward the benefit becomes clearer with larger head-only lifts, suggesting a threshold at which meta-gradients can provide reusable entity features without crowding out the LM signal.

\begin{figure}[!hp]
  \centering
  \includegraphics[width=\linewidth]{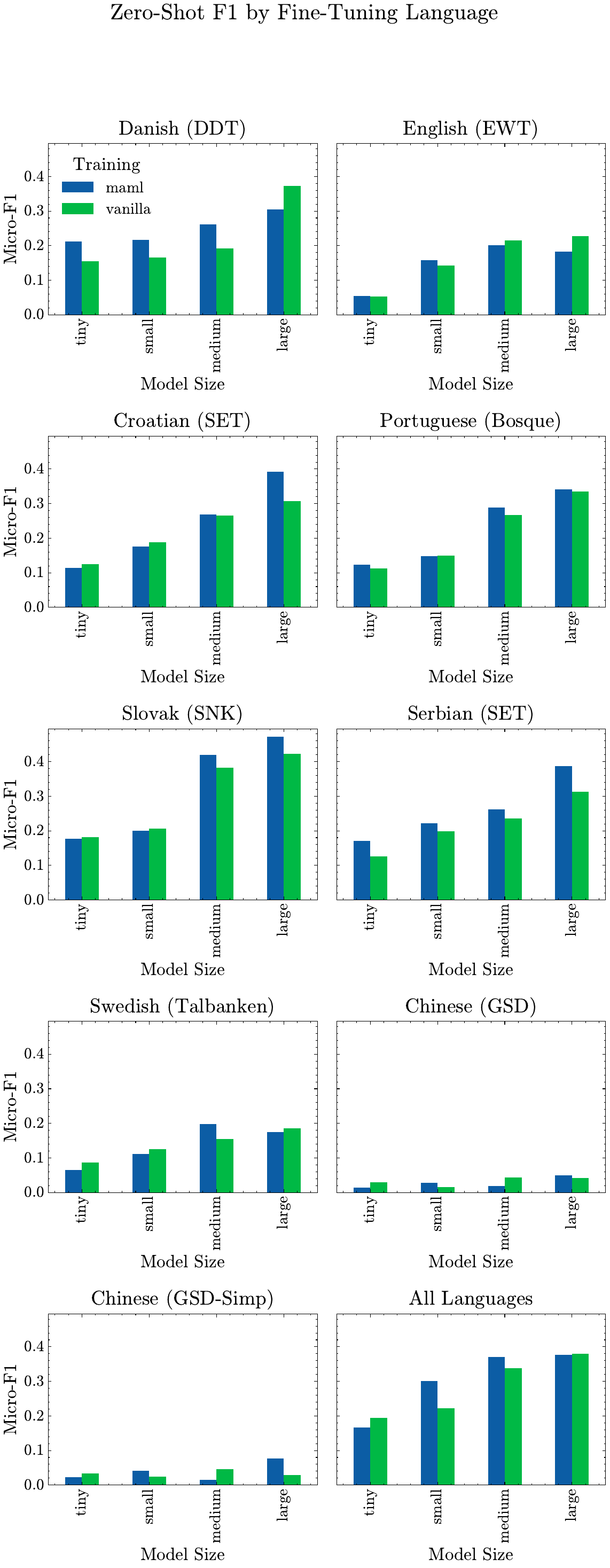}
  \caption{\textbf{Sensitivity to finetuning language.}  
           Grid of zero-shot F$_1$ curves after adapting on nine high-resource languages plus an \emph{All}-languages mixture.  
           Eight of nine languages show positive deltas; the largest relative gains occur for Slovak and Croatian, while Simplified Chinese is the lone outlier (–2 pp).  
           This pattern indicates that the meta-objective encourages reliance on surface affixes and particles that generalise well across Indo-European sources yet still transfer to Austronesian targets.}  \label{fig:zero_shot_by_lang}
\end{figure}

\paragraph{Sensitivity to finetuning language.} Figure \ref{fig:zero_shot_by_lang} profiles performance after adapting on nine high-resource languages. Eight of nine languages exhibit positive deltas; the largest relative lifts occur for Slovak (+18 \%) and Croatian (+13 \%). Gain in Slovak might be due to fixed case endings that consistently bracket entity names, providing a clear surface boundary signal for the model (similar in function to Tagalog’s case particles but realised morphologically rather than syntactically.) The sole regression (–2 pp on Simplified Chinese) is most likely due to a known issue in poor cross-script transfer to Chinese, but it may also be due to subword sparsity in the shared vocabulary rather than a failure of the meta-objective. \citep{mayhew-etal-2024-universal}.

Overall, MAML appears to teach the model to exploit shallow lexical anchors (particles, affixes) that generalise well across Indo-European languages while still transferring to more typologically distant Austronesian targets. To better understand the mechanisms underlying these gains, we conduct a focused qualitative analysis on a representative configuration.

\section{Analysis of MAML Pretrained Models}

In order to analyze the learning process, rather than just the last checkpoint, we focus our qualitative study on a \textsc{medium}-sized model (181 M parameters) finetuned in a head-only regime on Slovak (\texttt{sk\_snk}), finetuning on all 61 checkpoints from step 0 of pretraining to step 6000. We restrict our analysis to this slice because while finetuning 9760 (2 pretraining regimes x 2 finetuning regimes x 4 model sizes x 10 finetuning languages x 61 checkpoints) models would be prohibitively expensive, this configuration at least offers a reasonable signal-to-cost trade-off. This is for a few reasons: (i) the medium tier is the smallest model that still exhibits a clear 2–3 pp head-only lift (Figure \ref{fig:zero_shot_by_size}) yet is three-times cheaper to run than the 570 M variant, (ii) Slovak delivers one of the largest relative gains without vocabulary sparsity issues and, as a Slavic language, should produce transfer errors that differ sharply from those in Tagalog and Cebuano, and (iii) freezing the backbone during head-only finetuning ensures that any performance delta must stem from representations learned during meta-pretraining rather than from subsequent weight updates. In the next subsection, we inspect how pretraining affects finetuning performance across checkpoints.

\subsection{Checkpoint Analysis}

\begin{figure}[t]
  \centering
  \includegraphics[width=.9\linewidth]{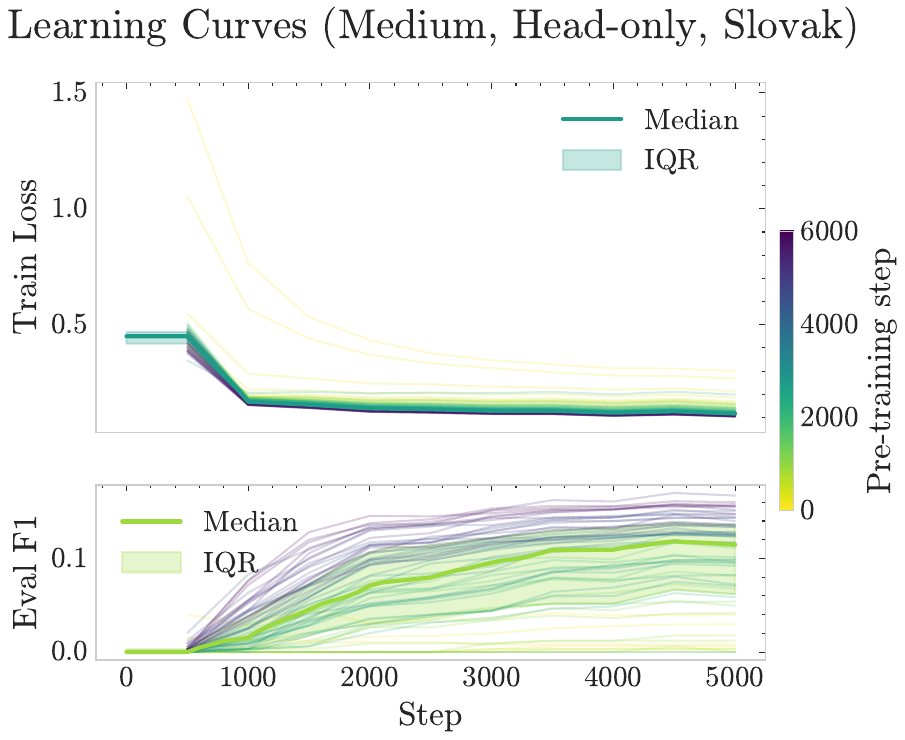}
  \caption{\textbf{Learning curves for the Slovak head-only setting.}
           Top: train loss; bottom: eval micro-F$_1$.
           Faint green lines = all individual checkpoints;
           bold line = median; shaded band = 25–75 \% IQR.
           Both metrics converge monotonically and remain tightly bunched,
           indicating a stable optimisation surface for the linear head.}
  \label{fig:sk_learning_curves}
\end{figure}

\paragraph{Does the head-only learner actually learn?}
Figure~\ref{fig:sk_learning_curves} overlays the complete finetuning trajectories for every Slovak head-only run
($61$ checkpoints, \texttt{maml\_s0000}–\texttt{maml\_s6000}). Viridis traces show the individual runs (getting darker the later the model checkpoint was taken), while the bold line and ribbon denote the median and inter-quartile range (IQR). The train-loss fan collapses to its asymptote within the first
$\approx\!800$~steps and stays flat thereafter; in parallel the evaluation F$_1$ rises smoothly to $0.14$ and plateaus with a narrow $\pm0.01$ IQR. Crucially, no run diverges or oscillates, confirming that
freezing the backbone and training only a linear chain CRF head is both stable and something is learned. This satisfies the prerequisite for using the configuration as a clean
test-bed: any downstream difference between MAML and vanilla is likely to stem
from the initial representations, not from optimisation quirks or training instabilities.

\begin{figure}[t]
  \centering
  \includegraphics[width=.95\linewidth]{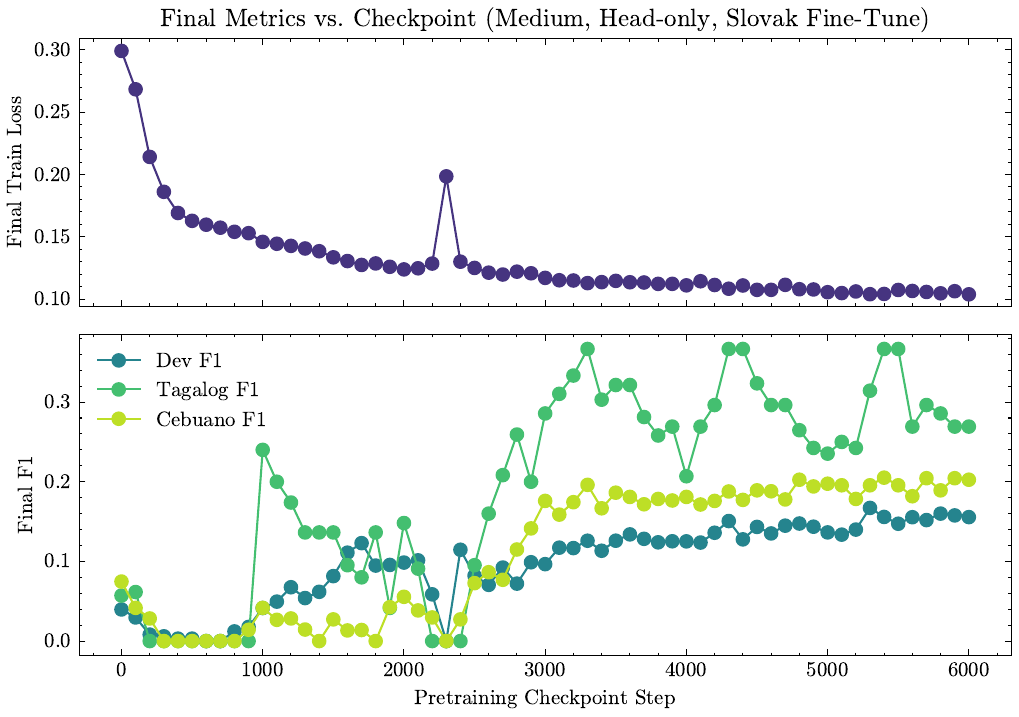}
  \caption{\textbf{Final metrics vs.\ pretraining checkpoint} for the
           \textsc{medium} MAML backbone frozen during head-only
           finetuning on Slovak. Top: final train loss of the
           CRF head, every run converges to the same narrow range.
           Bottom: final micro-F$_1$ on Slovak dev (blue),
           Tagalog (green) and Cebuano (yellow).  Although in-language
           performance saturates early, cross-lingual F$_1$ keeps
           improving up to step 6000, indicating that later meta-updates
           learn representations useful specifically for zero-shot
           transfer.}
  \label{fig:checkpoint_sweep}
\end{figure}

\begin{figure}[!htbp]
\centering
\includegraphics[width=.95\linewidth]{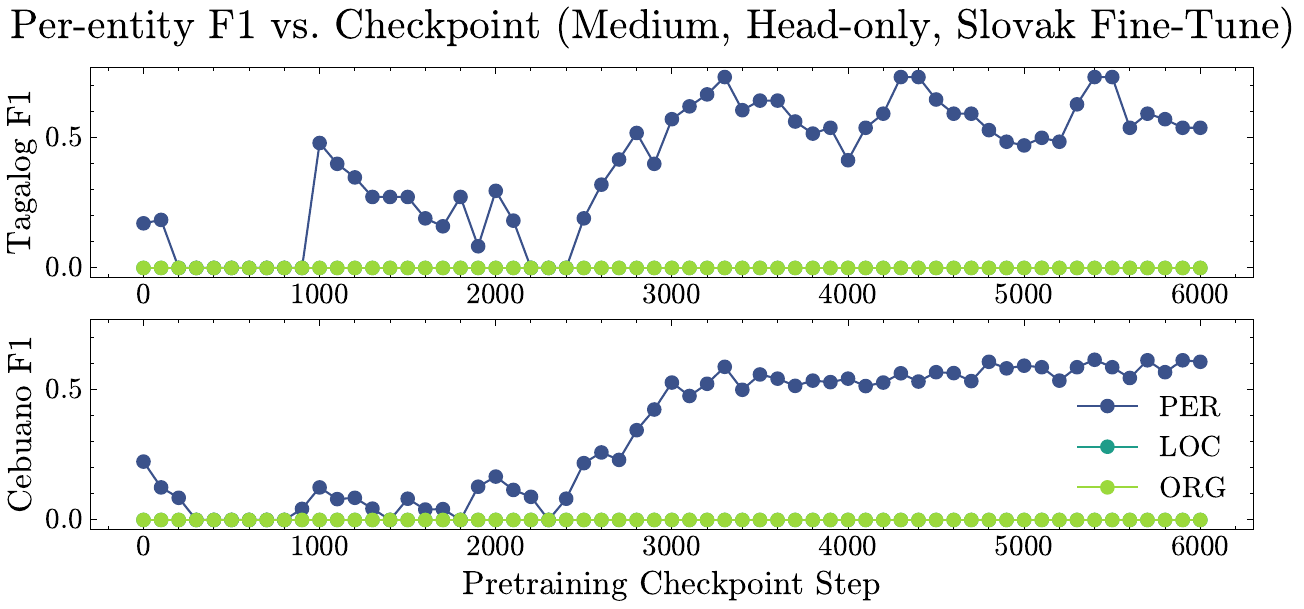}
\caption{\textbf{Per-entity F$_1$ across MAML checkpoints.}
PER (dark viridis) improves steadily with more meta-steps;
LOC and ORG curves remain at chance level, indicating that the frozen backbone provides transferable features for single-token personal names but little for multi-token locations or organisations.
Tagalog benefits earlier than Cebuano, consistent with its obligatory case particles.}
\label{fig:tagwise}
\end{figure}

\paragraph{Does meta-pretraining yield transfer-relevant representations?}
The checkpoint sweep in Figure \ref{fig:checkpoint_sweep} confirms the other prerequisite for this qualitative analysis: that meta-pretraining produces representations which become increasingly helpful for zero-shot transfer. First, the top panel shows that, regardless of which MAML snapshot we freeze, the linear chain CRF head always converges to essentially the same narrow band of train loss (0.10-0.15); optimisation is therefore stable and predictable, satisfying our first prerequisite. More importantly, the bottom panel reveals a very different story for cross-lingual evaluation: while Slovak dev F\textsubscript{1} plateaus early (by around step 1k), Tagalog and Cebuano F\textsubscript{1} continue to climb for another four thousand meta-updates, ending 0.15 and 0.12 points higher than at the initial checkpoint. In other words, additional MAML steps learn features that are invisible to the in-language dev set yet directly benefit unseen Austronesian targets. Tagalog improves earlier and peaks higher than Cebuano, hinting that the meta-objective is capturing surface cues (e.g.\ case particles) that are more diagnostic in Tagalog. Taken together with the “fan” plot of learning curves, the sweep demonstrates that meta-pretraining yields encoder states that are both optimisation-friendly and transfer-relevant, justifying the focus on this snapshot for deeper qualitative inspection. As such, we deepen the analysis in the next subsection by inspecting the behavior of our models on the level of the NER tags predicted.

\subsection{Tag-level Analysis}

\paragraph{Per-tag behaviour.}
Figure \ref{fig:tagwise} reports per-entity F\textsubscript{1} obtained after head-only finetuning the Slovak CRF head on each MAML checkpoint. \textsc{PER} climbs to 0.6-0.7 while \textsc{LOC} and \textsc{ORG} remain at zero. This is not a case of the classifier “over-fitting” in the usual sense—i.e.\ collapsing to always predicting a single label. A linear‐chain CRF is free to emit any BIO tag at any position; if it were truly degenerate we would see train loss stagnate near the log-uniform baseline and the PER curve itself would also be flat. Instead, train loss converges to the same narrow band for every checkpoint (Fig.\ref{fig:sk_learning_curves}) and PER performance tracks the amount of meta-pretraining, so the head is learning a genuine decision boundary. It simply has informative features for people but none for locations or organisations.

\paragraph{Observed imbalance and potential causes.}  First, the Slovak finetune set is intrinsically person-heavy.  As Table \ref{tab:sk_stats} shows, \textsc{PER} spans outnumber \textsc{LOC} by roughly $8{:}1$ and \textsc{ORG} by $15{:}1$.  Under head-only training, every gradient step passes through the frozen encoder unchanged and the CRF receives thousands of positive updates for persons but only a few hundred for the other classes. This likely leads to only the \textsc{PER} decision boundary sharpening.  Second, 87.6\% of Slovak person mentions are single tokens compared with 75.1 \% for locations and 56.9\% for organisations.  A single-token span can be captured by one weight vector, whereas multi-word spans require the head to model boundaries and label transitions—a capacity it simply does not have when the encoder cannot adapt.  Third, Tagalog still offers a comparatively reliable surface cue. The case particles \emph{si} and \emph{ni} precede roughly 11\% of gold \textsc{PER} spans, almost double the 5–6 \% rate observed in Cebuano (Table \ref{tab:particle_oov}). The earlier lift and higher ceiling of the Tagalog \textsc{PER} curve are therefore consistent with the backbone having learned to map the pattern "particle + token" to the \textsc{PER} label, a cue that is informative in Tagalog but is sparser in Cebuano.  Finally, cross-lingual lexical overlap is likely higher for personal names, many of which (e.g.\ \textit{Obama}, \textit{Manuel}) appear verbatim in English corpora used during pretraining; locations and organisations, by contrast, are often translated or abbreviated.  All four factors act in the same direction, favouring \textsc{PER}. Disentangling their individual contributions would require targeted ablations (particle masking, balanced resampling, controlled name substitution, etc.) which we leave for future work. In the next subsection, we assess behaviors on the level of words and tokens to relate NER performance to the low-resource languages being transferred to.

\subsection{Word-level Analysis}
\label{sec:wordlevel} 
\begin{figure*}[t]
  \centering

  \begin{subfigure}[b]{0.48\linewidth}
    \centering
    \includegraphics[width=\linewidth]{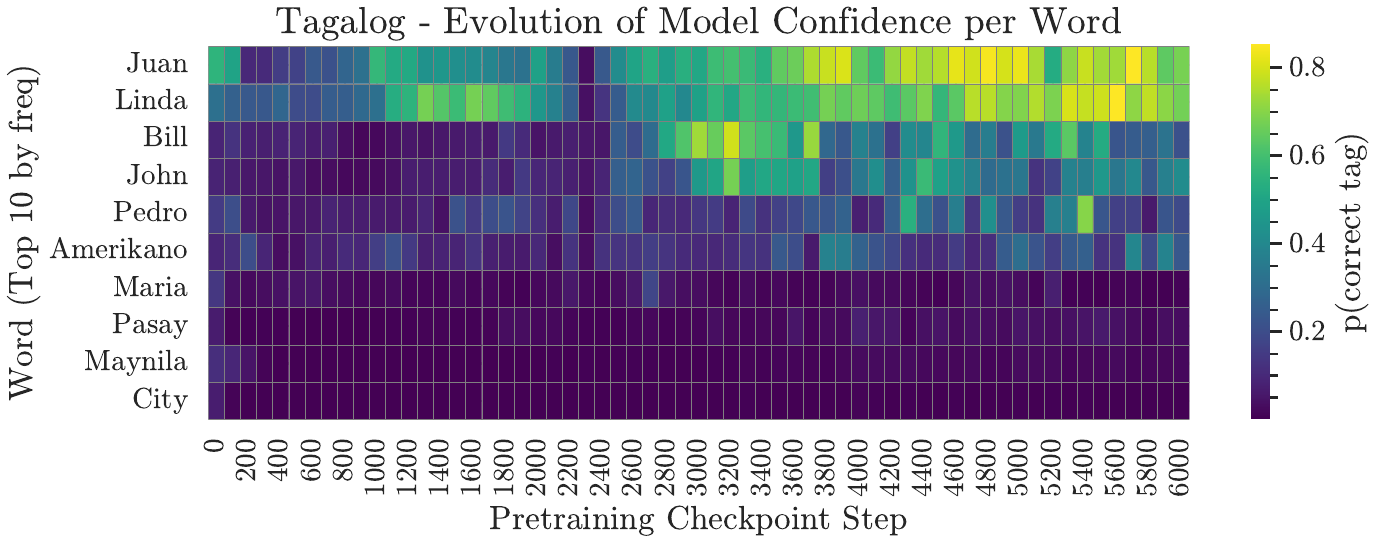}
    \caption{Tagalog — Entities only}
    \label{fig:tl_ent_heat}
  \end{subfigure}
  \hfill
  \begin{subfigure}[b]{0.48\linewidth}
    \centering
    \includegraphics[width=\linewidth]{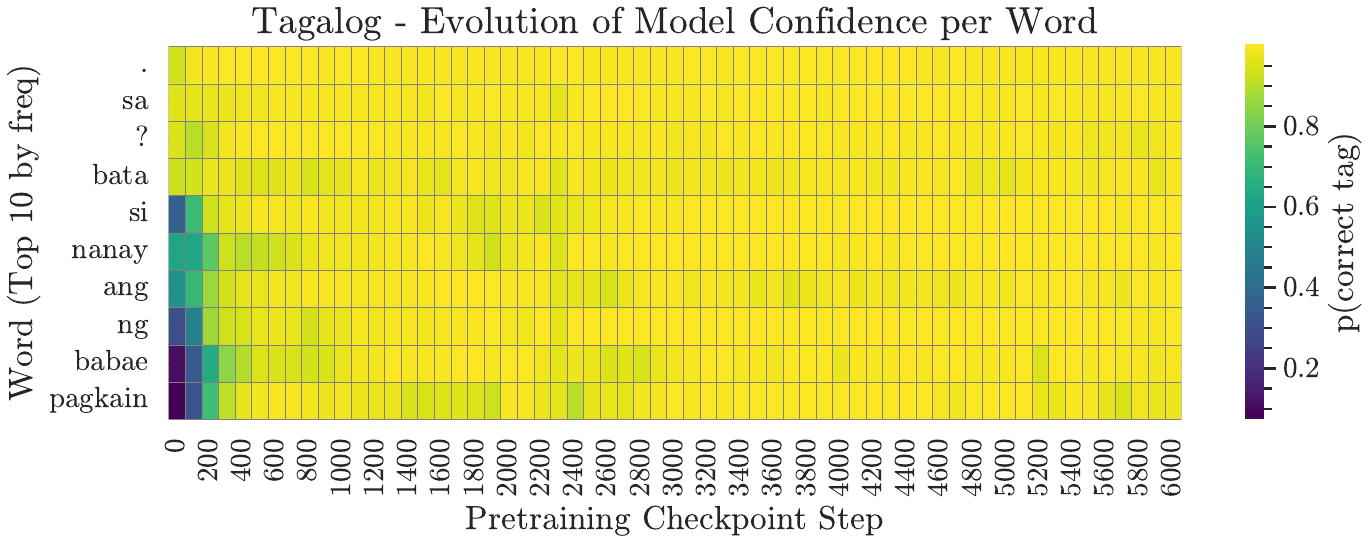}
    \caption{Tagalog — Non-entities}
    \label{fig:tl_o_heat}
  \end{subfigure}

  \vspace{0.9em} 

  \begin{subfigure}[b]{0.48\linewidth}
    \centering
    \includegraphics[width=\linewidth]{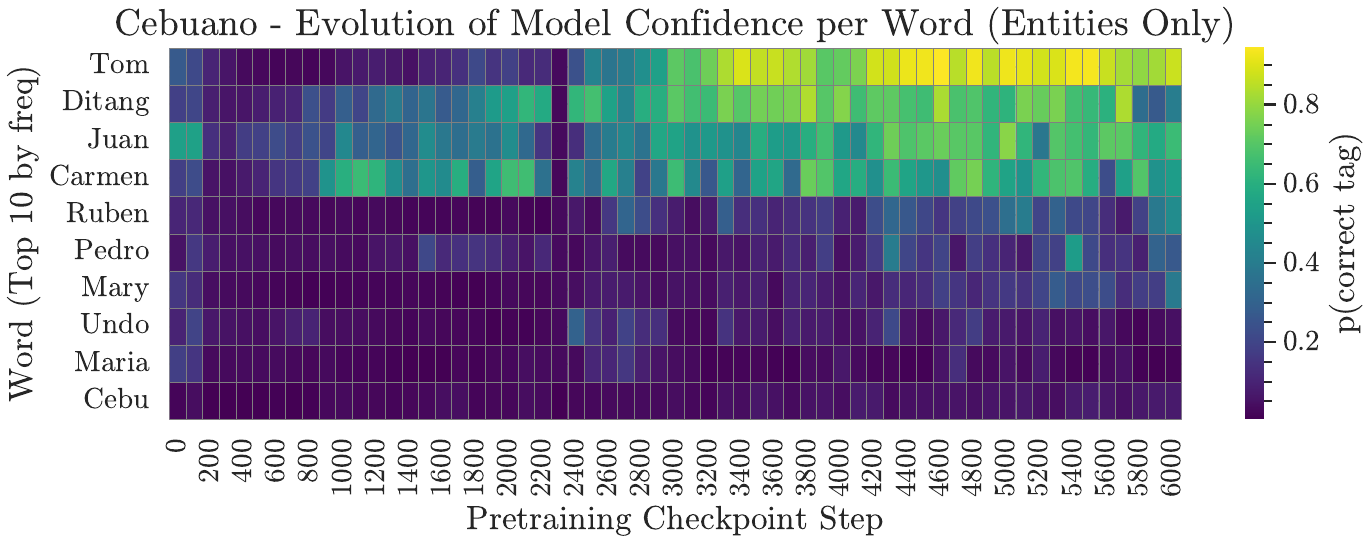}
    \caption{Cebuano — Entities only}
    \label{fig:ceb_ent_heat}
  \end{subfigure}
  \hfill
  \begin{subfigure}[b]{0.48\linewidth}
    \centering
    \includegraphics[width=\linewidth]{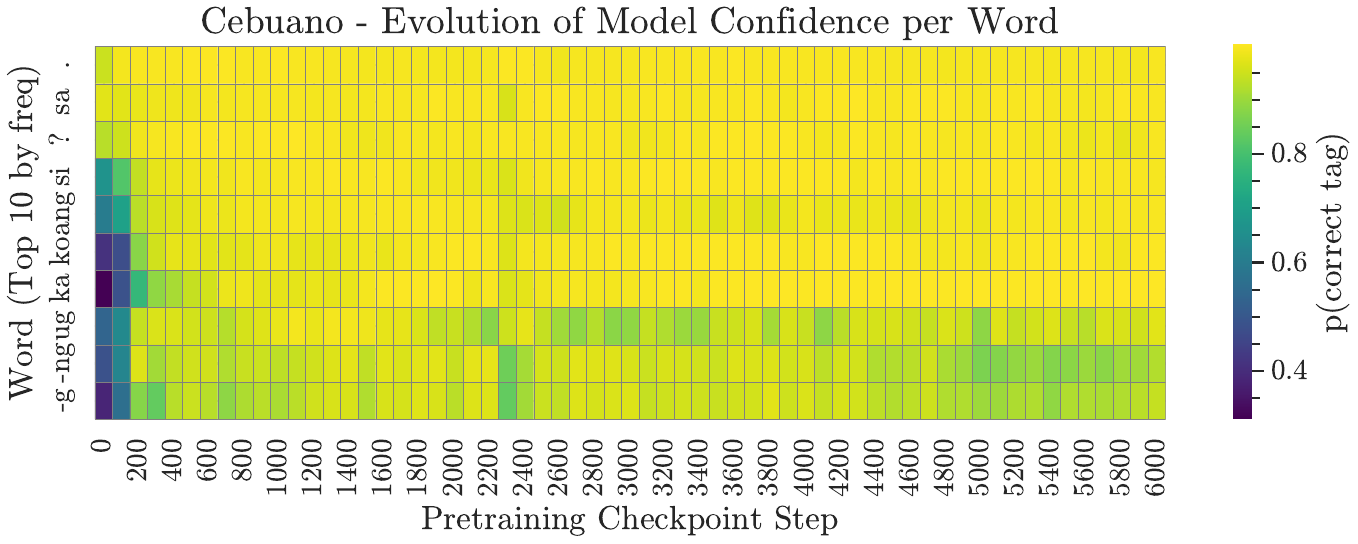}
    \caption{Cebuano — Non-entities}
    \label{fig:ceb_o_heat}
  \end{subfigure}

  \caption{\textbf{Evolution of token-level confidence} ($p(\text{correct tag})$) across pretraining checkpoints.
  Top row: Tagalog; bottom row: Cebuano. Left: entities only. Right: non-entities.}
  \label{fig:all_heatmaps}
\end{figure*}

Figures~\ref{fig:tl_ent_heat}–\ref{fig:ceb_o_heat} visualise the
checkpoint‐by‐checkpoint evolution of token-level confidence ($p(\text{correct tag})$) for the ten most frequent surface words in each evaluation set. Entities and non-entities are split so the
dynamic range is not drowned out by \textsc{O} tokens.
Two qualitative patterns emerge.

\paragraph{Fast confidence in frequent tokens.}
Non-entity function words such as \emph{ng, ang, sa} in Tagalog and the
Cebuano clitic \emph{-ng} start with high confidence and barely
budge after the first \(200\) meta-updates
(Fig.~\ref{fig:tl_o_heat},~\ref{fig:ceb_o_heat}).  As these tokens dominate
the language-model loss, autoregressive training achieves a high confidence in them
early and MAML has little head-room to improve over checkpoints.

\paragraph{Monotonic gains for high-overlap proper names.}
In the Tagalog set, international names (\emph{City, Maynila, Maria}) and
locations transliterated from English (\emph{Pasay}) become steadily
brighter (lower loss) until about step~\(3000\)
(Fig.~\ref{fig:tl_ent_heat}).  Similar behaviour appears for
\emph{Maria, Cebu, Mary} in Cebuano
(Fig.~\ref{fig:ceb_ent_heat}).  These words either appear verbatim in
the English Dolma corpus or share sub-tokens (\texttt{Ma\_}, \texttt{Ceb\_}) with it, so the
meta-objective can reuse prototypes that happen to be used by the
Austronesian targets.  The timing matches the checkpoint-sweep
(Fig.~\ref{fig:checkpoint_sweep}): cross-lingual F\textsubscript{1}
continues to climb long after Slovak dev has saturated likely because the
back-bone is still lowering loss on these anchor words. We illustrate these mechanisms further in two case studies in Appendix \ref{app:casestudies}.



\section{Finetuning Speed of Meta-Pretraining}

Finally, we assess finetuning speed using convergence time (measuring time to achieve 90\% of final loss $t_{90}$), normalized area under the loss curve (measuring aggregate convergence behavior over the curve), and initial slope (measuring the initial speed of learning in the first few steps), as seen in Table \ref{tab:finetune_speed_deltas}. Across nine in-language tasks, full-model finetuning shows the clearest acceleration for the largest and smallest models: MAML cuts $t_{90}$ by roughly 8\% ($\approx 111$ steps) and modestly reduces loss AUC. Medium and small models show negligible or inconsistent speed-ups under full tuning, suggesting that the effect depends strongly on model capacity. In head-only tuning, large and small models again benefit slightly, while medium and tiny models slow down, likely due to underpowered or collapsed meta-dynamics.

\begin{table}[!t]
\centering
\small
\begin{tabular}{llrrr}
\toprule
\textbf{Size} & \textbf{Regime} & $\Delta t_{90}$ & $\Delta\mathrm{AUC}$ & $\Delta\mathrm{slope}$ \\
\midrule
\texttt{large}  & full   & -111.1 & -0.004 &  0.0e-05 \\
                & head   &  -55.6 & -0.012 &  1.0e-05 \\
\addlinespace
\texttt{medium} & full   &    0.0 & -0.005 &  0.0e-05 \\
                & head   &   55.6 & -0.011 &  0.0e-05 \\
\addlinespace
\texttt{small}  & full   &    0.0 &  0.003 &  0.0e-05 \\
                & head   &  -55.6 &  0.003 & -0.0e-05 \\
\addlinespace
\texttt{tiny}   & full   & -111.1 &  0.004 & -0.0e-05 \\
                & head   &   55.6 & -0.023 &  5.0e-05 \\
\bottomrule
\end{tabular}
\caption{Finetuning convergence speed metrics $\Delta$ (MAML-Vanilla) averaged over nine in-language tasks. The largest and smallest models enjoy the most pronounced speed-ups from full MAML meta-initialization, while medium and tiny models show negligible $\Delta t_{90}$ under full-model tuning. Under head-only tuning, large and small decoders still benefit modestly, whereas medium and tiny decoders actually slow down. Across all settings, slope remains near zero, indicating that meta-training primarily accelerates mid-to-late convergence rather than the very first gradient steps.}
\label{tab:finetune_speed_deltas}
\end{table}

Initial slopes remain effectively unchanged across all settings, indicating that MAML does not alter the very first gradient steps but instead reorganizes the loss landscape to make mid- to late-stage convergence more efficient. These results align with earlier findings that MAML’s main benefit lies in providing sharper, more reusable token-level features for high-capacity backbones, with limited or negative effects when capacity is insufficient to retain both language modeling and episodic priors.

\section{Related Work}

\paragraph{NER in Filipino, Tagalog, and Cebuano.}
NER for Philippine languages remains underexplored, with most work focusing on resource construction rather than cross-lingual modeling. Recent corpora include TLUnified-NER \citep{miranda-2023-developing}, TF-NERD \citep{ramos2023tf},  CebuaNER \citep{pilar-etal-2023-cebuaner}, and UniversalNER \citep{mayhew-etal-2024-universal}. Modeling efforts in this area primarily use NER-specific systems \citep{10.1145/3719384.3719444, ebona2013named, 8650208} incorporating a simpler backbone such as a support vector machine \citep{castillo2013named} or an LSTM \citep{chan2023practical}. Most recently, FilBench \citep{miranda2025filbenchllmsunderstandgenerate} and Batayan \citep{montalan-etal-2025-batayan} support Filipino evaluation on NLP tasks for LLMs.

\paragraph{Meta-learning for Pretraining.}
Although most work applies meta‐learning at fine‐tuning time, a growing line of research embeds meta‐objectives directly into pretraining. \cite{raghu2021meta} showed that framing parameter-efficient adapter learning as a bilevel problem yields representations that fine‐tune more effectively than standard PEFT. \cite{hou-etal-2022-meta} extend this to full transformers.  \cite{miranda2023pretrainingtrulybettermetalearning} argue that explicit MAML objectives can outperform fixed pretraining on highly diverse task distributions. \cite{ke-etal-2021-pre} integrate a MAML‐style inner loop into a multi-criteria Chinese Word Segmentation pretraining task.
\section{Conclusion}

This paper shows that MAML-based meta-pretraining, even when applied to small decoder-only language models, can meaningfully improve zero-shot transfer to low-resource languages, as demonstrated on Tagalog and Cebuano NER. The gains are most pronounced for person entities and head-only finetuning, and scale best with larger model capacities. Our qualitative and word-level analyses reveal that the mechanism of improvement centers on the sharpening of lexical prototypes and better anchoring to surface cues like Tagalog case particles. Hence, we do not expect these improvements to fully generalize to multi-token or highly contextual entity types.

These findings suggest that meta-learning can provide a principled route to more adaptable small models, but also highlight key limitations: the benefits are capacity- and task-dependent, and the current approach struggles with richer entity structures. Future work should explore alternative meta-learning objectives, extend to more diverse tasks and languages, and investigate the dynamics of prototype formation in even lower-resource settings. 

\section*{Limitations}
The gains are most pronounced for person entities and head-only finetuning, and scale best with larger model capacities. All training runs stop at exactly six thousand outer steps, a horizon that may be too short for the largest model, so the conclusions derived only cover a fraction of the training budget a corporate setup might have. A more diverse and multilingual corpus may alter both quantitative and qualitative conclusions, and varying languages in the meta-task is a natural way to extend this work. Qualitative analysis was conducted on a single configuration and single seed due to cost and GPU constraints. Qualitative analysis was conducted by a native Tagalog speaker with a register typical of Manila, and a wide variety of perspectives would improve the robustness of the analysis. Finally (and most naturally), our focus on only two Austronesian languages controls for certain lexical and syntactic divergences but limits the generality of the typological conclusions; extending to a broader set of Philippine and Malayo-Polynesian languages is a natural next step.

\bibliography{latex/acl_latex}

\clearpage
\appendix
\onecolumn 
\section{NER-Relevant Typological Features of Cebuano and Tagalog  }
\label{app:linguistic}
This extended table highlights how morphosyntactic and discourse-level differences between the two languages interact with the challenges of named entity recognition (NER). We lay out feature-by-feature contrasts to illustrate that even closely related Philippine languages present distinct hurdles for tasks like NER. The table emphasizes that while Tagalog offers overt morphosyntactic cues (e.g., case particles, topic marking), Cebuano relies more heavily on discourse inference, thereby requiring different modeling strategies for effective NER.

\begin{table*}[!ht]
\centering
\small
\begin{tabular}{p{3cm}p{3cm}p{3cm}p{7cm}}
\toprule
\textbf{Typological Feature} & \textbf{Tagalog} & \textbf{Cebuano} & \textbf{Challenge for NER} \\
\midrule
Voice system & Four-way actor/non-actor voice paradigm & Reduced two-way system & Tagalog’s rich voice alternations encode argument roles morphologically, complicating alignment of entities with semantic roles. Cebuano’s reduced system lowers redundancy, making cues for role identification less explicit. \\
Case marking & Obligatory case particles (si, ni, ang, ng, sa) & Case particles often dropped or fused & Tagalog provides reliable morphosyntactic signals for entity boundaries/roles. Cebuano forces reliance on discourse, requiring coreference and contextual inference. \\
Lexical borrowing / code-switching & High density of Spanish loans and English code-switching & More conservative Austronesian lexicon & Tagalog NER must cope with OOV issues, language-mixing, and orthographic variation. Cebuano NER must handle morphologically complex Austronesian stems, underrepresented in multilingual embeddings. \\
Morphological richness & Productive affixation (focus, aspect, causatives) & Similarly rich, but slightly more regular & Surface forms for named entities may be inflected or derivationally complex, increasing sparsity for training data. \\
Word order flexibility & Relatively free (voice and particles constrain roles) & Even freer, especially without explicit case markers & Named entities may appear in non-canonical positions, reducing the utility of positional cues. \\
Pronominal systems & Rich system of clitic pronouns that attach to verbs or particles & Similar system but with different distributions & Entities can be referred to obliquely or dropped entirely; clitic attachment blurs tokenization boundaries, confusing NER pipelines. \\
Reduplication & Common for aspect, plurality, intensification & Widespread and productive & Reduplicated forms of named entities (nicknames, reduplicated roots) may not be recognized as related to the canonical form. \\
Orthography \& variation & Spanish-influenced orthography, multiple spelling conventions & More phonologically consistent, but dialectal spelling variation persists & Orthographic inconsistency makes lexicon-based NER brittle, especially in noisy social media text. \\
Discourse prominence / topic marking & Ang-marked topic influences salience & Topic is often inferred from discourse, less explicit marking & Tagalog gives overt topic marking, aiding salience detection; Cebuano relies on pragmatics, requiring discourse-level modeling. \\
\bottomrule
\end{tabular}
\caption{Detailed typological contrasts between Tagalog and Cebuano and their implications for NER.}
\label{tab:tagalog-cebuano-appendix}
\end{table*}

\begin{figure}[!hb] \small \setlength{\tabcolsep}{4pt} 
\begin{tabular}{llllllll} \emph{Tag.} & Pumunta & si & \textbf{Maria} & sa & \textbf{Cebu}.\\ Gloss & go.\textsc{pfv} & \textsc{nom} & Maria & \textsc{obl} & Cebu\\ NER & O & O & B-PER & O & B-LOC\\[4pt]

\emph{Ceb.} (with marker) & Miadto & si & \textbf{Juan} & sa & \textbf{Sugbo}. \\ Gloss & go.\textsc{pst} & \textsc{nom} & Juan & \textsc{obl} & Cebu \\ NER & O & O & B-PER & O & B-LOC \\[2pt]

\emph{Ceb.} (zero-marked) & Miadto & \textbf{Juan} & sa & \textbf{Sugbo}. \\ Gloss & go.\textsc{pst} & Juan & \textsc{obl} & Cebu \\ NER & O & B-PER & O & B-LOC \\ \end{tabular} \caption{Surface cues for named entities. Tagalog typically provides an overt personal article (\emph{si}/\emph{ni}) before names; Cebuano may show the same article, but zero‑marked variants also occur in some registers/contexts, reducing overt anchors.} \label{fig:intro-example} \end{figure}

\newpage 

\section{Case Studies}
\label{app:casestudies}
To illustrate the mechanisms underlying MAML's improvements, we present two contrasting examples that demonstrate how meta-pretraining affects different types of linguistic patterns in Tagalog NER. We measure $\Delta$ log-prob as the change in surprisal ($-!\log p$) for the gold label between the vanilla and MAML model. A negative $\Delta$ means the model is more confident after MAML; a positive $\Delta$ means less confident.

\paragraph{Case 1: Prototype Amplification.}
Sentence: “Inahit ni John ang sarili niya.” (Gloss: “John shaved himself.”)

The first case study demonstrates how MAML strengthens recognition of cross-linguistically common proper names. In this example, MAML sharply reduces surprisal on “John,” indicating stronger prototype activation.

We suspect improvement operates at two levels: (1) lexical level, in the sense that the token "John" becomes more strongly associated with person entities through meta-learning's emphasis on rapid adaptation to new entities, and (2) contextual level, in the sense that the \emph{ni} + proper-name pattern gets reinforced as a reliable \textsc{PER} indicator during meta-training episodes.

\paragraph{Case 2: Contextual Suppression (Loss).}
Sentence: “Malapit kay Maria si Juan.” (Gloss: “Juan is close to Maria.”)

The second case study reveals MAML's limitations with complex multi-token constructions. Here, $\Delta$ is positive for key tokens, showing that MAML reduces confidence in the correct label. In "Malapit kay Maria si Juan" (Juan is close to Maria), both the locative adverb "Malapit" (close/near) and the oblique case marker "kay" show substantially decreased confidence for location labeling under MAML (combined decrease of approximately $-3.3$ log-probability points).

We suspect this occurs due to: (1) capacity constraints, in the sense that the frozen backbone has limited representational capacity, and strengthening \textsc{PER} features may crowd out \textsc{LOC}/\textsc{ORG} representations, and (2) training signal imbalance, in the sense that finetuning contained more person-like entities than complex locative expressions, biasing the learned representations toward single-token person recognition.

\begin{figure}[!htbp]
  \centering
  \begin{subfigure}[t]{0.49\textwidth}
    \includegraphics[width=\linewidth]{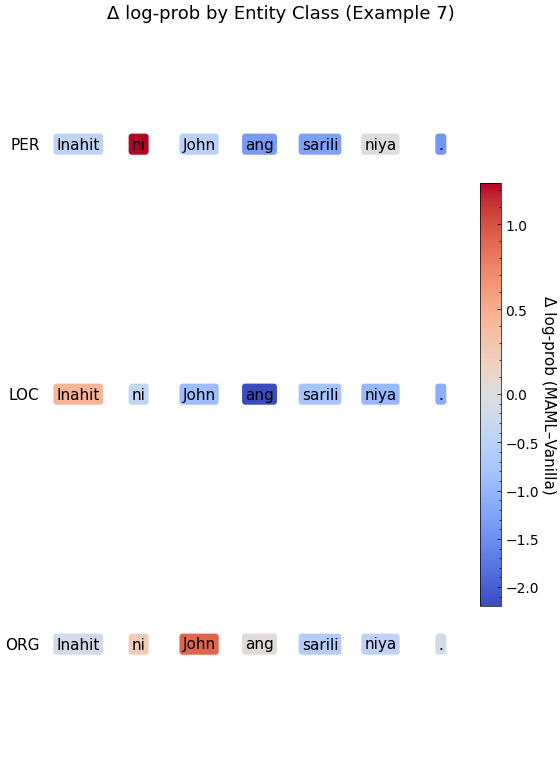}
    \caption{Prototype Amplification.}
  \end{subfigure}
  \hfill
  \begin{subfigure}[t]{0.49\textwidth}
    \includegraphics[width=\linewidth]{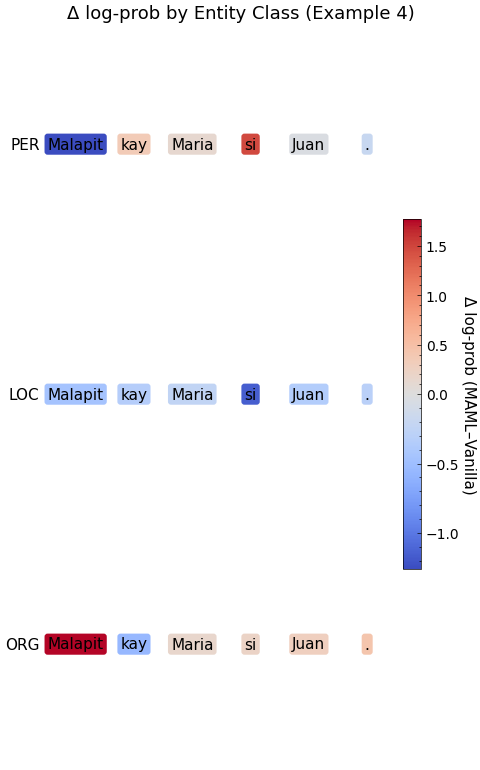}
    \caption{Contextual Suppression.}
  \end{subfigure}
  \caption{MAML's impact on (a) single-token prototype confidence and (b) multi-token contextual cue sensitivity.}
  \label{fig:case_studies}
\end{figure}

\newpage 
\section{Pseudocode}
\label{app:pseudocode}
Below is the pseudocode for the MAML and vanilla pretraining setup.

\subsection*{Distributed Subset Masked Language Modeling Tasks (SMLMT) Training}
\begin{algorithm}[H]
\caption{Distributed SMLMT Loop}\label{alg:dist-maml}
\begin{algorithmic}[1]
  \State // \emph{Initialization:} same as Alg.~\ref{alg:dist-ar}, plus
  \State initialize inner‐optimizer SGD on head $h_\phi$
  \State step $\leftarrow 0$

  \For{each sub\_batch in dataloader}
    \State // gather across GPUs
    \State $X \gets$ \texttt{fabric.all\_gather}(sub\_batch["input\_ids"])
    \State // sync random branch decision
    \State $r \gets \text{Uniform}(0,1)$; $r \gets$ \texttt{fabric.broadcast}($r$)
    \If{$r < \rho$} 
      \State // \emph{Meta‐learning episode}
      \State $(S,Q),\,\mathrm{labels}_S,\mathrm{labels}_Q \gets \mathrm{mask\_tokens}(X)$
      \State $\phi_0\gets\phi$  \Comment{snapshot head params}
      \For{$t=1$ \textbf{to} $T_{\textrm inner}$}
        \State $\ell_S\gets\mathrm{CE}(h_{\phi_{t-1}}(f_\theta(S)),\mathrm{labels}_S)$
        \State $\phi_t\gets\phi_{t-1}-\alpha\,\nabla\ell_S$  \Comment{inner SGD}
      \EndFor
      \State $\ell_Q\gets\mathrm{CE}(h_{\phi_T}(f_\theta(Q)),\mathrm{labels}_Q)$
      \State $\phi\gets\phi_0$  \Comment{restore head}
      \State \texttt{fabric.backward}($\ell_Q$/accum\_steps)
    \Else
      \State // \emph{Standard AR}
      \State $X_{\text{in}},Y\gets X[\,:\!,:-1\,],\,X[\,:\!,1:\,]$
      \State $\ell_{\text{AR}}\gets\mathrm{CE}(f_\theta(X_{\text{in}}),Y)$
      \State \texttt{fabric.backward}($\ell_{\text{AR}}$/accum\_steps)
    \EndIf

    \State // outer‐step and logging
    \If{(step+1)\,\%\,accum\_steps == 0}
      \State opt.step(); scheduler.step(); opt.zero\_grad()
      \State // aggregate metrics across GPUs
      \State log\_loss $\gets$ \texttt{fabric.all\_reduce}($\ell$)
      \State \texttt{fabric.log}(\dots)
      \State \texttt{fabric.barrier}()
    \EndIf

    \State step $\mathrel{+}{=}$ 1
  \EndFor
\end{algorithmic}
\end{algorithm}

\subsection*{Distributed Autoregressive (AR) Training}
\begin{algorithm}[H]
\caption{Distributed AR Loop}\label{alg:dist-ar}
\begin{algorithmic}[1]
  \State // \emph{Initialization (in \texttt{Trainer.\_\_init\_\_}):}
  \State Load configs; initialize Fabric, tokenizer, model $f_\theta$
  \State $(\text{model}, \text{opt}) \gets \texttt{fabric.setup}(f_\theta,\,\mathrm{AdamW})$
  \State dl $\gets$ base dataloader; dl $\gets$ \texttt{fabric.setup\_dataloaders}(dl)
  \State step $\leftarrow$ 0; zero gradients

  \For{each sub\_batch in dl}
    \State // \emph{Gather full batch across GPUs if needed:}
    \State $X \gets$ \texttt{fabric.all\_gather}(sub\_batch["input\_ids"])
    \State $X_{\text{in}}, Y \gets X[\,:\!,:-1\,],\,X[\,:\!,1:\,]$
    \State // forward + loss
    \State $\ell \gets \mathrm{CE}\bigl(f_\theta(X_{\text{in}}),\,Y\bigr)$
    \State // backward (handles synchronization)
    \State \texttt{fabric.backward}$\bigl(\ell\text{/accum\_steps}\bigr)$
    \State // outer‐step when accumulated
    \If{(step+1)\,\%\,accum\_steps == 0}
      \State opt.step(); scheduler.step(); opt.zero\_grad()
      \State // optional barrier
      \State \texttt{fabric.barrier}()
    \EndIf
    \State step $\mathrel{+}{=}$ 1
  \EndFor
\end{algorithmic}
\end{algorithm}

\subsection{Multi-GPU processing}
Pico already uses Lightning-Fabric data parallelism but meta-learning introduces various demands that make multi-GPU processing complicated. A Bernoulli draw is done on one GPU and broadcast so all ranks choose the same objective. Support and query tensors are constructed on rank 0 then scattered, because per-rank random masks would destroy gradient equivalence. Every GPU performs the same ten head updates before any gradient is communicated.  A stray early \texttt{all\_reduce} would mix gradients from different inner steps, so we place an explicit \texttt{barrier} between inner and outer phases.

\section{Universal NER Datasets}
\label{app:uner}

To comprehensively evaluate the pretraining method, each permutation of finetuning setup (\{head-only, full\}, finetuning dataset (\{\texttt{da\_ddt}, \dots, \texttt{zh\_gsdsimp}, \texttt{all}\}) (where \texttt{all} consists of all available training sets), model size (\{tiny, small, medium, large\}), and pretraining setup (\{vanilla, MAML\}) is evaluated, for a total of 160 evaluation runs.

\begin{itemize}
  \item \textbf{Publicly Available In-language treebanks} (9 langs): full \texttt{train}/\texttt{dev}/\texttt{test} splits, identical to the official UD partitions.  
    \begin{itemize}
      \item \texttt{da\_ddt}, \texttt{en\_ewt}, \texttt{hr\_set}, \texttt{pt\_bosque}, \texttt{sk\_snk}, \texttt{sr\_set}, \texttt{sv\_talbanken}, \texttt{zh\_gsd}, \texttt{zh\_gsdsimp}
    \end{itemize}
  \item \textbf{Parallel UD (PUD) evaluation} (6 langs): single \texttt{test.txt} files, all sentence-aligned across German, English, Portuguese, Russian, Swedish and Chinese.  
    \begin{itemize}
      \item \texttt{de\_pud}, \texttt{en\_pud}, \texttt{pt\_pud}, \texttt{ru\_pud}, \texttt{sv\_pud}, \texttt{zh\_pud}
    \end{itemize}
  \item \textbf{Other eval-only sets} (3 langs): small test splits for low-resource languages.  
    \begin{itemize}
      \item \texttt{ceb\_gja} (Cebuano), \texttt{tl\_trg} (Tagalog TRG), \texttt{tl\_ugnayan} (Tagalog Ugnayan)
    \end{itemize}
\end{itemize}
\newpage 
\subsection{Slovak Fine-Tune Token Statistics}

\begin{table}[!ht]
\centering
\begin{tabular}{lrrr}
\toprule
\textbf{Entity} & \textbf{\# spans} & \textbf{\% single-token} \\
\midrule
PER & 2\,277 & 87.6\,\% \\
LOC &   277 & 75.1\,\% \\
ORG &   153 & 56.9\,\% \\
\bottomrule
\end{tabular}
\caption{Span statistics for the Slovak finetune set
(\texttt{sk\_snk}\,train).  The data are strongly person-heavy and
person spans are almost always single words, whereas locations and
organisations are both rarer and more often multi-token.}
\label{tab:sk_stats}
\end{table}

\subsection{Tagalog and Cebuano Particle and Out-of-Vocabulary Statistics}

\begin{table}[!ht]
\centering
\begin{tabular}{lcc}
\toprule
\textbf{Language} & \textbf{Particle recall} & \textbf{OOV rate} \\
\midrule
Tagalog & 0.113 $\;\pm\;$ 0.000 & 0.523 $\;\pm\;$ 0.000 \\
Cebuano & 0.058 $\;\pm\;$ 0.000 & 0.534 $\;\pm\;$ 0.000 \\
\bottomrule
\end{tabular}
\caption{Mean (\,$\pm$\,s.d.\ across checkpoints) of particle–preceding-span
recall and token out-of-vocabulary rate, measured on the zero-shot
evaluation sets after Slovak head-only tuning.  “Particle recall” is the
fraction of gold \textsc{PER} entities whose left context token is a Filipino
case particle recognised by the model.}
\label{tab:particle_oov}
\end{table}

\newpage 
\section{Pretraining Results}
\label{app:pretraining}
We present the unedited pretraining indicators for each \texttt{pico-maml-decoder} model below, as logged on WandB.

\begin{figure}[htbp]
  \centering
  \includegraphics[width=\textwidth]{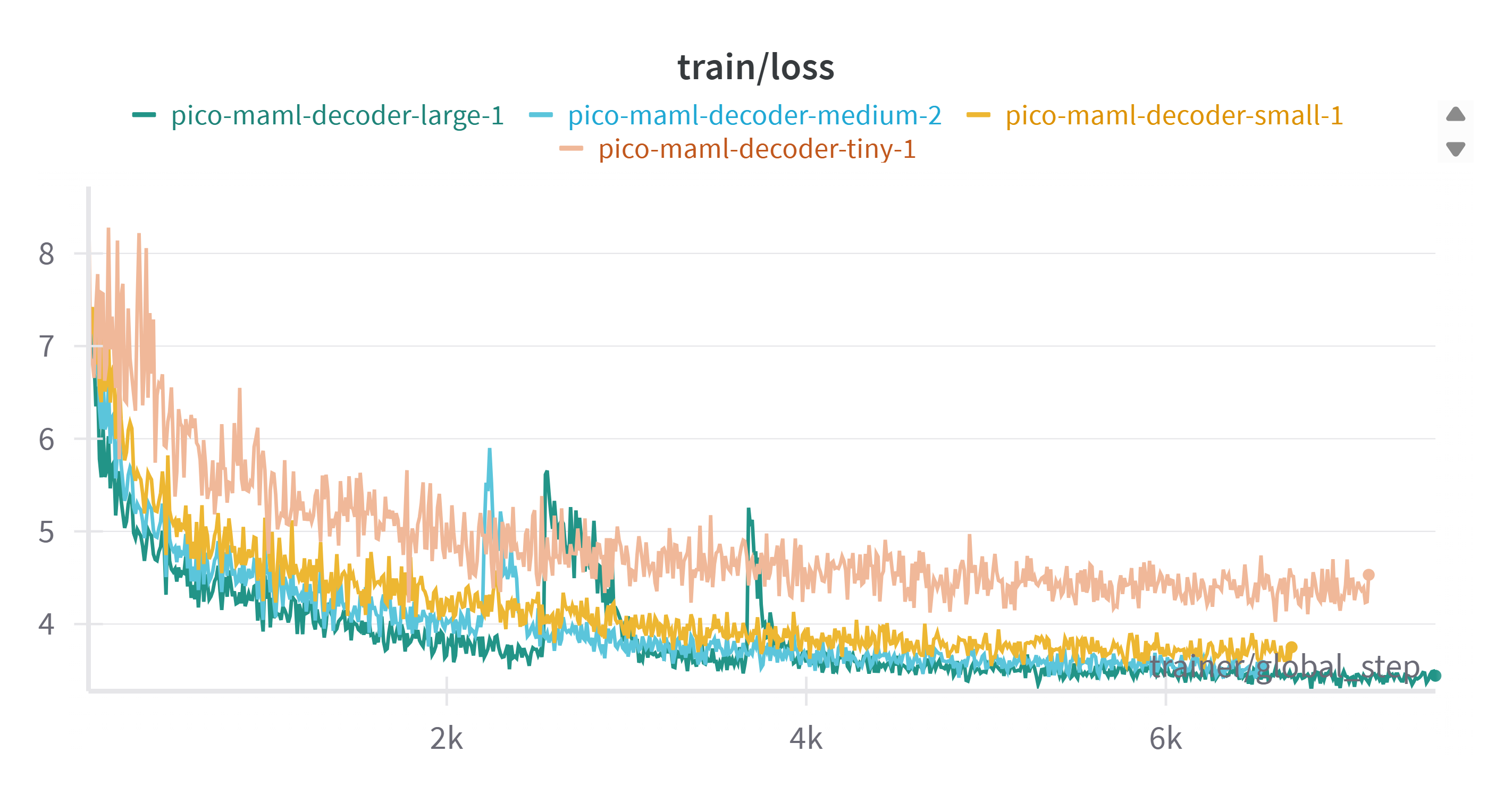}
  \caption{Pretraining training loss curve.}
  \label{fig:pretrain-loss}
\end{figure}

\begin{figure}[htbp]
  \centering
  \includegraphics[width=\textwidth]{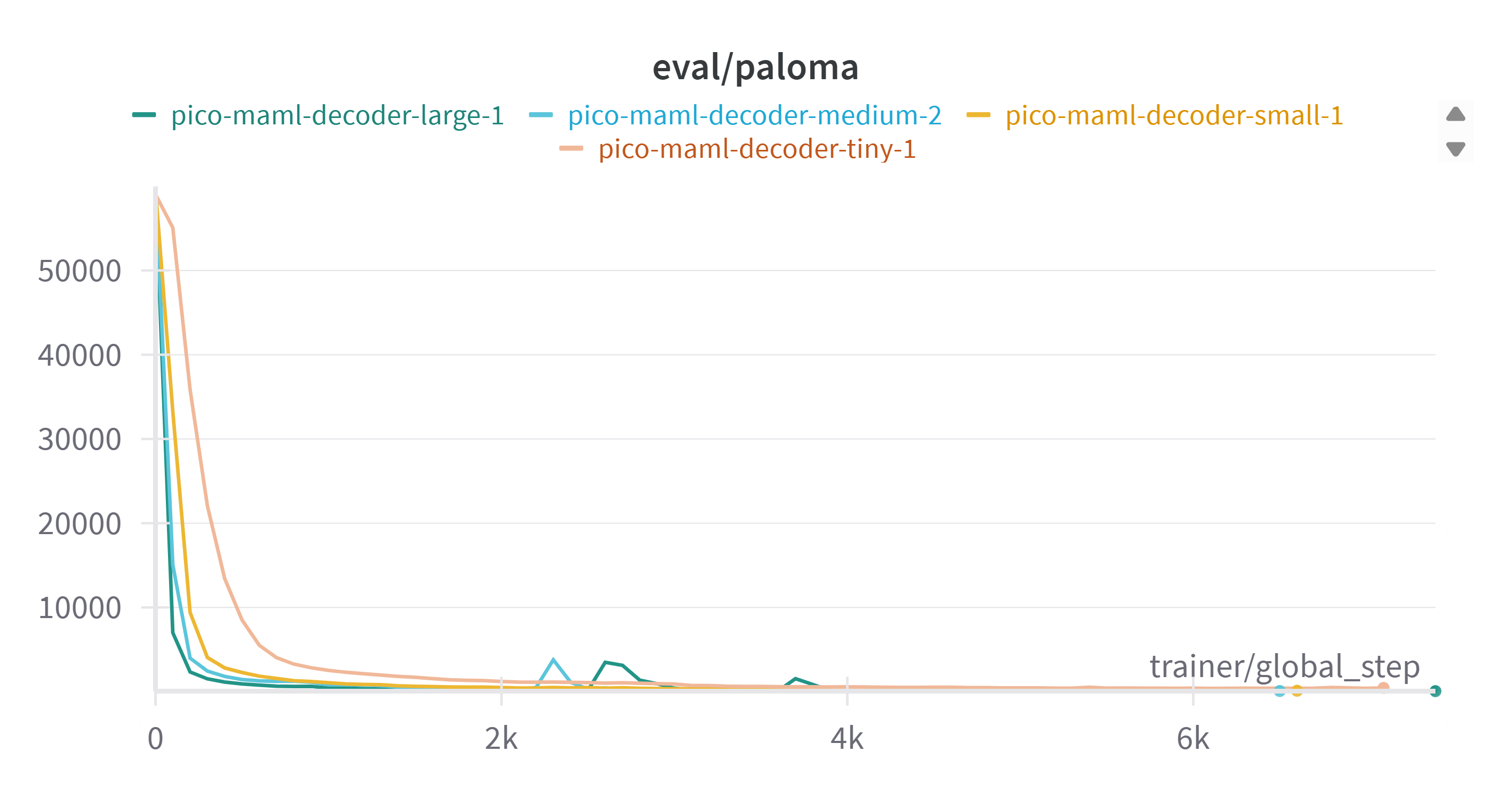}
  \caption{PALOMA score over pretraining steps.}
  \label{fig:pretrain-paloma}
\end{figure}

\begin{figure}[htbp]
  \centering
  \includegraphics[width=\textwidth]{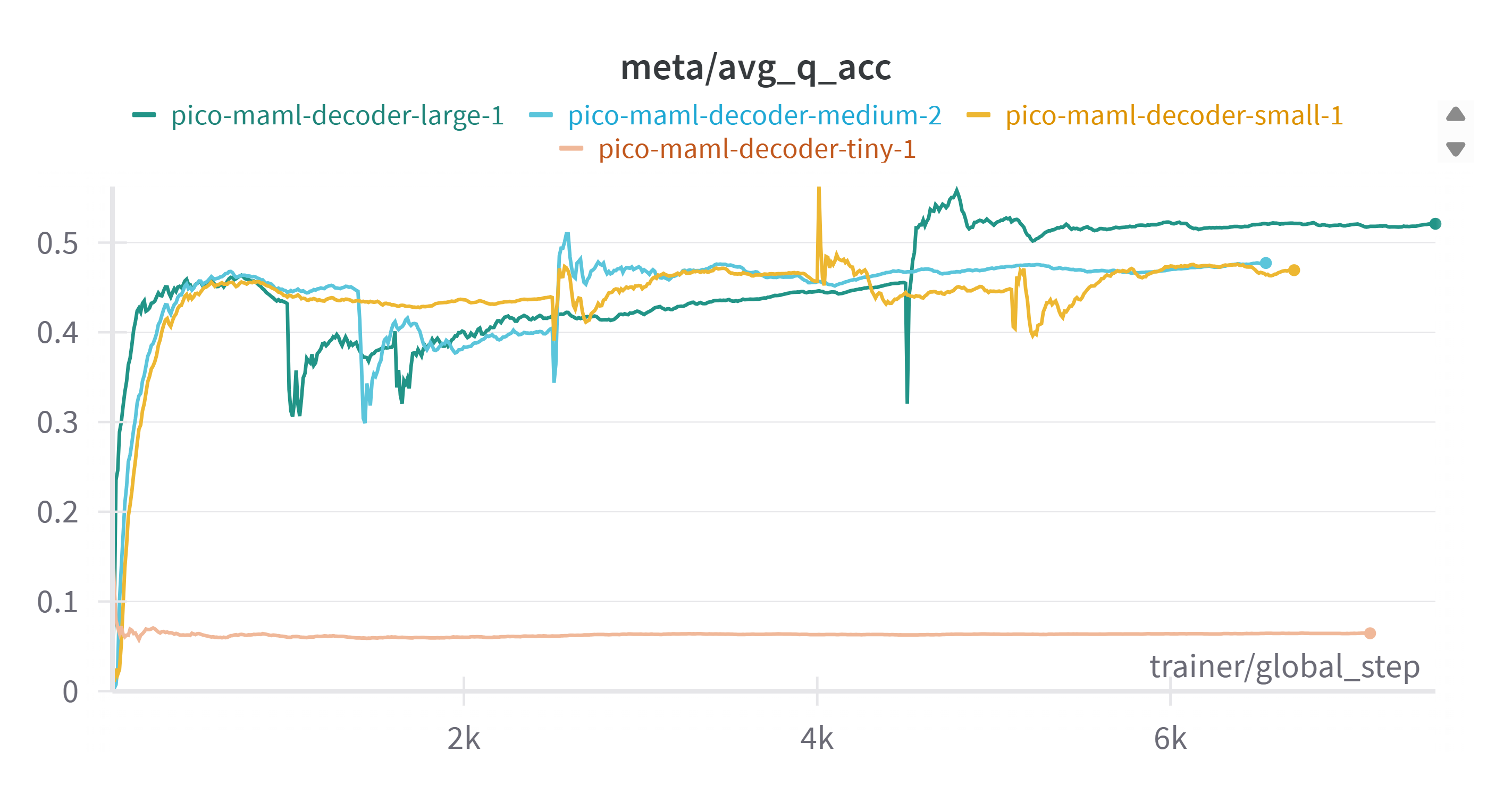}
  \caption{Query accuracy during pretraining.}
  \label{fig:pretrain-qacc}
\end{figure}

\begin{figure}[htbp]
  \centering
  \includegraphics[width=\textwidth]{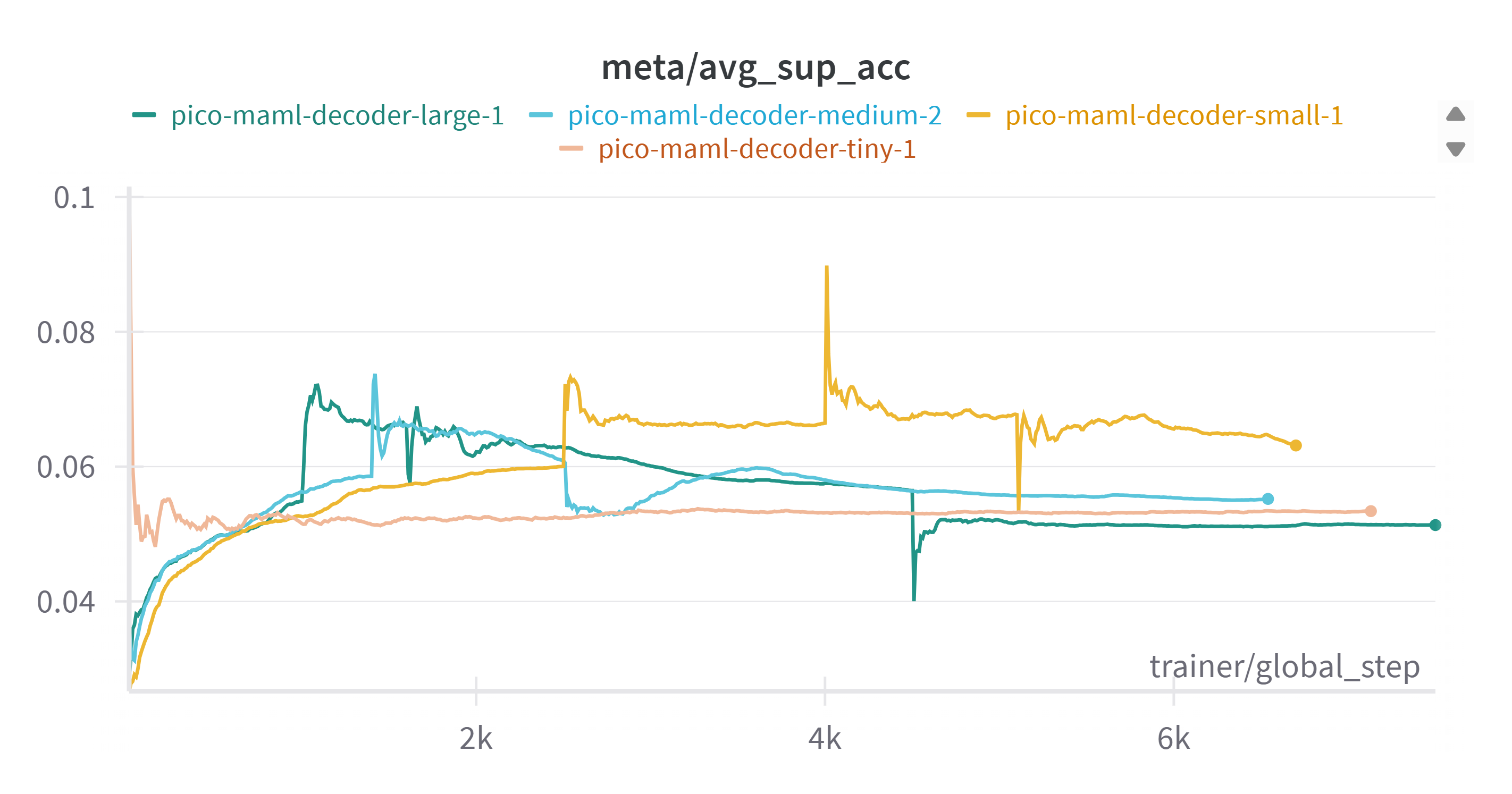}
  \caption{Support accuracy over pretraining.}
  \label{fig:pretrain-supacc}
\end{figure}

\begin{figure}[htbp]
  \centering
  \includegraphics[width=\textwidth]{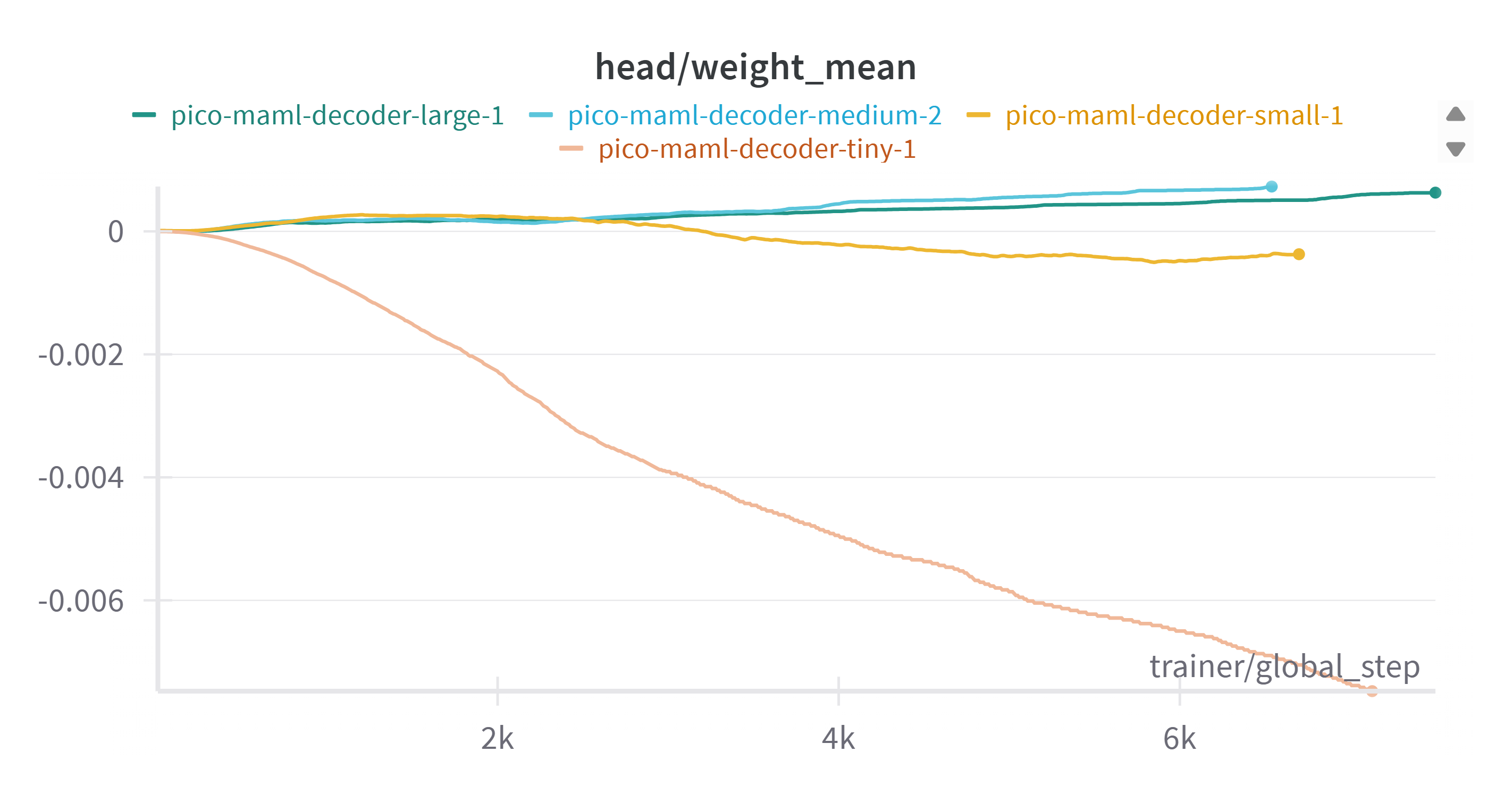}
  \caption{Mean of weights in classifier head over pretraining.}
  \label{fig:pretrain-wtmean}
\end{figure}

\begin{figure}[htbp]
  \centering
  \includegraphics[width=\textwidth]{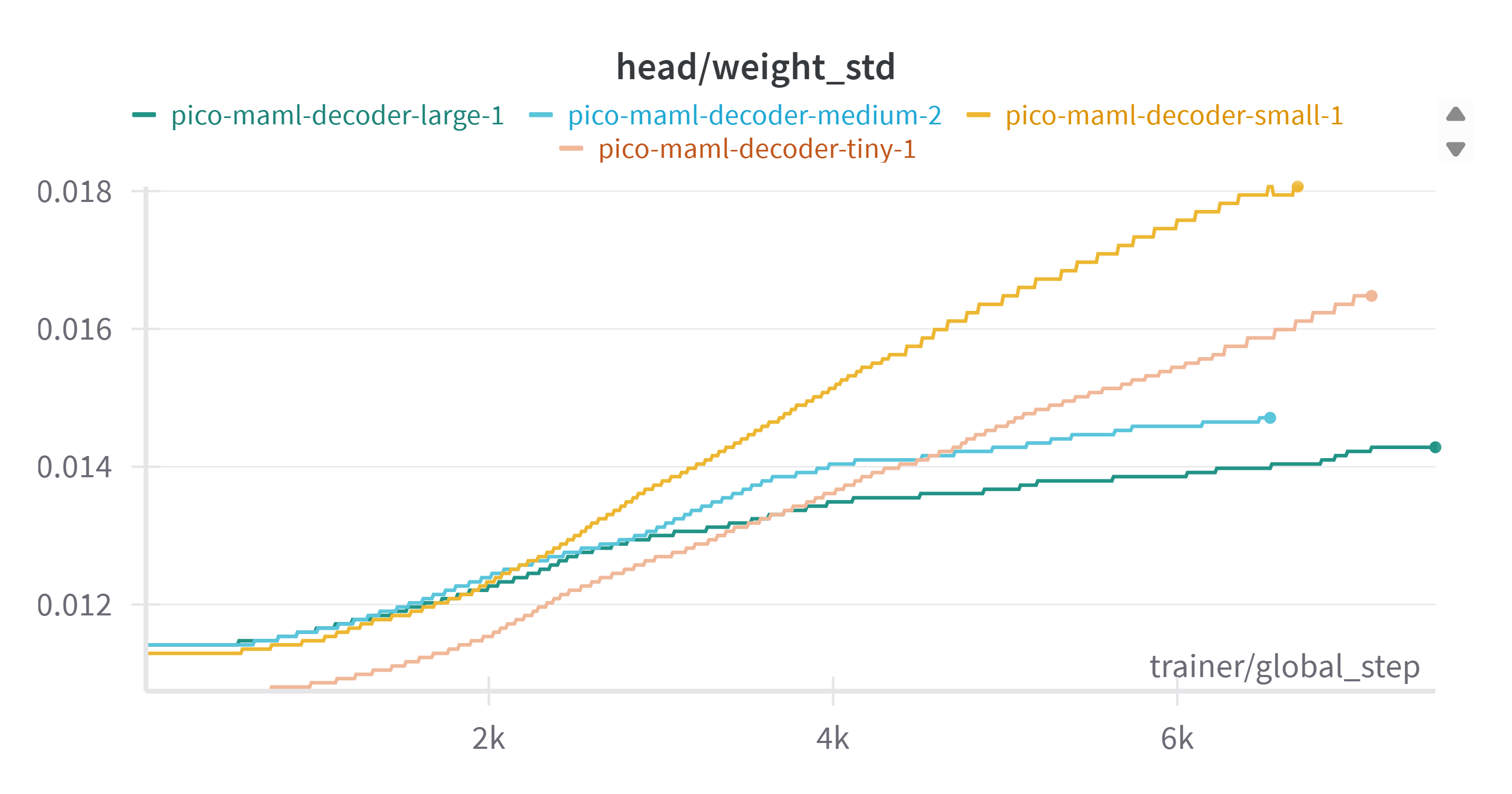}
  \caption{Standard deviation of weights in classifier head over pretraining.}
  \label{fig:pretrain-wtstd}
\end{figure}

\newpage
\section{Default \texttt{pico-maml-train} Configurations}
\label{app:config}
\begin{table*}[ht!]
    \centering
    \renewcommand{\arraystretch}{1} 
    \setlength{\tabcolsep}{8pt} 
    \begin{tabular}{|>{\centering\arraybackslash}p{3cm}|p{5cm}|p{6.5cm}|}
        \hline
        \textbf{Category} & \textbf{Parameter} & \textbf{Default Value} \\
        \hline
        \multirow{10}{*}{\textbf{Model}}  
            & Model Type & \texttt{pico\_decoder} \\
            & Hidden Dimension ($d_{\text{model}}$) & 768 \\
            & Number of Layers ($n_{\text{layers}}$) & 12 \\
            & Vocabulary Size & 50,304 \\
            & Sequence Length & 2,048 \\
            & Attention Heads & 12 \\
            & Key/Value Heads & 4 \\
            & Activation Hidden Dim & 3,072 \\
            & Normalization Epsilon & $1 \times 10^{-6}$ \\
            & Positional Embedding Theta & 10,000.0 \\
        \hline
        \multirow{7}{*}{\textbf{Training}}  
            & Optimizer & AdamW \\
            & Learning Rate & $3 \times 10^{-4}$ \\
            & LR Scheduler & Linear w/ Warmup \\
            & Warmup Steps & 2,500 \\
            & Gradient Accumulation Steps & 128 \\
            & Max Training Steps & 200,000 \\
            & Precision & BF16 Mixed \\
        \hline
        \multirow{3}{*}{\textbf{Data}}  
            & Dataset Name & \texttt{pico-lm/pretokenized-dolma} \\
            & Batch Size & 1,024 \\
            & Tokenizer & \texttt{allenai/OLMo-7B-0724-hf} \\
        \hline
        \multirow{6}{*}{\textbf{Checkpointing}}  
            & Auto Resume & True \\
            & Save Every N Steps & 100 \\
            & Learning Dynamics Layers & \texttt{"attention.v\_proj",} \newline \texttt{"attention.o\_proj",} \newline \texttt{"swiglu.w\_2"} \\
            & Learning Dynamics Eval Data & \texttt{pico-lm/pretokenized-paloma-tinsy} \\
        \hline
        \multirow{3}{*}{\textbf{Evaluation}}  
            & Metrics & \texttt{["paloma"]} \\
            & Paloma Dataset Name & \texttt{pico-lm/pretokenized-paloma-tinsy} \\
            & Eval Batch Size & 16 \\
        \hline
        \multirow{3}{*}{\textbf{Monitoring}}  
            & Logging Level & INFO \\
            & Log Every N Steps & 100 \\
        \hline
                \multirow{12}{*}{\textbf{Meta-Learning}}  
            & Enabled & True \\
            & Hybrid Ratio & 0.5 \\
            & Inner Steps ($k$) & 10 \\
            & Inner Learning Rate & 0.001 \\
            & Support Shots ($k$) & 4 \\
            & Query Ways ($n$) & 32 \\
            & Classifier Head Layers & 4 \\
            & Classifier Head Hidden Dim & 128 \\
            & Classifier Head Dropout & 0.1 \\
            & Classifier Head Init Method & \texttt{xavier} \\
        \hline
        \multirow{3}{*}{\textbf{Monitoring}}  
            & Logging Level & INFO \\
            & Log Every N Steps & 100 \\
        \hline
    \end{tabular}
    \caption{Default configuration settings used in \texttt{pico-maml-train}.}
    \label{tab:defaultconfig}
\end{table*}

\newpage 
\begin{table*}[!]
\centering
\renewcommand{\arraystretch}{1}
\begin{tabular}{|p{0.30\textwidth}||p{0.13\textwidth}|p{0.13\textwidth}|p{0.13\textwidth}|p{0.13\textwidth}|}
\hline
\multicolumn{5}{|c|}{\textbf{Pico-MAML-Decoder Model Comparison}} \\
\hline
\textbf{Attribute} & \texttt{tiny} & \texttt{small} & \texttt{medium} & \texttt{large} \\
\hline
Parameter Count & 11M & 65M & 181M & 570M \\
Hidden Dimension ($d_{\text{model}}$) & 96 & 384 & 768 & 1536 \\
Feed-forward Dim & 384 & 1536 & 3072 & 6144 \\
Training Time (6k steps) & 10h & 15h & 16h & 25h \\
\hline
\end{tabular}
\vspace{0.5em}
\caption{Comparison of \texttt{pico-maml-decoder} model variants trained with default \texttt{pico-maml-train} configurations. Except for hidden and feed-forward dimension, all models share the training settings detailed in \ref{tab:defaultconfig}. Models were trained for 6000 training steps on 4 NVIDIA A100-SXM4-80GB GPUs; the listed training times correspond to the initial 6000 steps.}
\label{tab:pico-maml-decoder-configs}
\end{table*}

\end{document}